\documentclass[11pt, a4paper, logo, copyright]{googledeepmind}

\usepackage[authoryear, sort&compress, round]{natbib}
\usepackage{multirow}
\usepackage{makecell}
\usepackage{adjustbox}
\usepackage{wrapfig}
\usepackage{float}
\usepackage{placeins}
\usepackage{pdflscape}
\usepackage{afterpage}
\usepackage{subcaption} 
\usepackage{svg}
\usepackage[super]{nth}

\usepackage{tabularray}
\usepackage{nicematrix}
\usepackage{longtable}
\usepackage{array}
\usepackage{enumitem}

\usepackage{bbm}
\usepackage{gensymb}
\usepackage{siunitx}
\usepackage{algorithm}
\usepackage{algorithmic}
\usepackage{xspace}
\usepackage{comment}
\usepackage{minted} 
\usepackage{tcolorbox}

\usepackage{tikz}
\usetikzlibrary{patterns, shapes.geometric, positioning}

\usepackage{minitoc}
\usepackage[toc,page,header]{appendix}
\usepackage{titletoc}

\usepackage{cleveref}

\newtcolorbox{surveybox}[1]{
    colback=gray!5, colframe=gray!50, colbacktitle=gray!30,
    coltitle=black, fonttitle=\bfseries, title={#1},
    arc=2mm, boxrule=0.5mm, left=6mm, right=6mm, top=4mm, bottom=4mm
}

\newlist{checklist}{itemize}{2}
\setlist[checklist]{label=$\square$}

\definecolor{bgcode}{rgb}{0.95,0.95,0.95}
\definecolor{textgreen}{RGB}{57, 172, 57}
\definecolor{textred}{RGB}{200, 10, 10}
\colorlet{darkgreen}{green!65!black}
\definecolor{baselinecolor}{gray}{.95}

\title{Uneven Evolution of Cognition Across Generations of Generative AI Models}
\correspondingauthor{$\{$jedm, isaacgl$\}$@google.com}
\reportnumber{} 
\renewcommand{\today}{2026-04-09}

\author[$\dagger$,1]{Isaac Galatzer-Levy}
\author[2,3]{Daniel McDuff}
\author[2,3]{Xin Liu}
\author[$\dagger$,2,3]{Jed McGiffin}

\affil[$\dagger$]{Corresponding Author}  
\affil[1]{Google DeepMind}
\affil[2]{Google Research}
\affil[3]{University of Washington}

\begin{document}

\begin{abstract}

The pursuit of artificial general intelligence necessitates robust methods for evaluating the cognitive capabilities of models beyond narrow task performance. Here, we introduce a psychometric framework to assess the cognitive profiles of generative AI, comparing them to human norms and tracking their evolution across generations. Initial evaluation of leading multimodal models using tasks adapted from the Wechsler Adult Intelligence Scale revealed a profoundly uneven cognitive architecture: near-ceiling performance in verbal comprehension and working memory (>\nth{98} percentile) contrasted with near-floor performance in perceptual reasoning (<\nth{1} percentile). To track developmental trajectories beyond human-normed limits, we developed the Artificial Intelligence Quotient (AIQ) Benchmark and applied it to six generations and two model families, revealing significant but asymmetric performance gains. Notably, we uncovered a sharp dissociation between modalities; abstract quantitative reasoning matured far more rapidly when presented linguistically compared to a visually analogous format, indicating an architectural bias towards language-based symbolic manipulation. While abstract visual reasoning improved, visual-perceptual organization remained largely stagnant. Collectively, these findings demonstrate that the cognitive abilities of generative models are evolving unevenly, suggesting that scaling and optimization approaches to AGI development alone may be insufficient to overcome fundamental architectural limitations in achieving balanced, human-like general intelligence.

\vspace{-0.1cm}
\end{abstract}

\maketitle

\setlength{\leftmargini}{9pt}

\vspace{0.3cm}
\section{Introduction}
\vspace{0.1cm}
\label{sec:intro}

The development of artificial general intelligence (AGI) requires evolving generative models from narrow experts towards systems capable of adaptive learning, reasoning, and novel problem-solving across diverse contexts \citep{goertzel2014artificial,legg2007collection}. Achieving this requires moving beyond pattern recognition to capabilities that mirror biological intelligence, which is dependent on core cognitive capabilities to perform highly diverse tasks \citep{lake2017building}. Standardized cognitive assessments, exemplified by the Wechsler Adult Intelligence Scale-Fourth Edition (WAIS-IV), provide the canonical framework for this type of evaluation in humans, offering normative metrics across established domains like verbal comprehension, perceptual reasoning, and working memory \citep{wechsler2008wechsler}. Their ability to reveal relative cognitive strengths and weaknesses makes them an ideal starting point for mapping the cognitive profile of AI.

To rigorously benchmark the cognitive development of AI, we employed a two-pronged psychometric approach. First, we adapted subtests from the WAIS-IV for administration to a range of contemporary multimodal models. This initial assessment revealed a starkly uneven cognitive profile: models demonstrated superlative performance at or above the \nth{98} percentile on Verbal Comprehension and Working Memory tasks, while exhibiting profound deficits in Perceptual Reasoning, with scores consistently below the \nth{1} percentile. This finding confirmed that while human-normed tests establish a critical baseline, their pronounced ceiling and floor effects render them insufficient for tracking the maturation of advanced AI capabilities.

Recognizing these limitations, we then developed the Artificial Intelligence Quotient (AIQ) Benchmark. This research protocol consists of novel datasets analogous to canonical cognitive tests but engineered to scale in difficulty far beyond human limits. By normalizing these instruments against the distribution of model performance, the AIQ provides a sensitive measure to overcome the ceiling effects observed with human tests. This dual methodology provides a comprehensive framework for charting the uneven evolution of cognition in generative models and identifying the architectural bottlenecks that may constrain progress toward AGI.
\section{Methods}
\label{sec:methods}

Our two-pronged approach involved adapting standardized human cognitive tests for AI model administration followed by the development of a novel AI-centric assessment battery, the AIQ Benchmark. 

\subsection{Adaptation of Standardized Cognitive Assessments}
\label{subsec:adapt_standard}

Widely considered the gold standard in clinical assessment, the Wechsler Adult Intelligence Scale (WAIS) is the most frequently used measure of cognitive functioning in the world (\citep{wright2017assessment}, designed to derive an overall intelligence quotient (IQ) by systematically sampling performance across core cognitive domains: verbal comprehension and reasoning, visual-perceptual reasoning, working memory, and processing speed. We initially adapted select subtests from the Wechsler Adult Intelligence Scale-Fourth Edition (WAIS-IV; \citep{wechsler2008wechsler}, targeting the Verbal Comprehension Index (VCI; Similarities, Vocabulary, Information, Comprehension), the Perceptual Reasoning Index (PRI; Matrix Reasoning, Visual Puzzles, Figure Weights, Picture Completion), and the Working Memory Index (WMI; Digit Span, Arithmetic, Letter-Number Sequencing). Subtests that required graphomotor capabilities (i.e. Block Design, Coding, Cancellation) were omitted resulting in an inability to calculate scores for the Processing Speed Index (PSI).

This initial phase involved administering the tests to several early- and current-generation multimodal models to establish a broad comparative baseline. The tested models included OpenAI's GPT-4 Turbo and GPT-4o, Google's Gemini Flash 1.5 and Pro 1.5, and Anthropic's Claude 3 Opus and Claude 3.5 Sonnet. WAIS-IV items were converted into text-based or multimodal prompts. Model outputs were scored according to WAIS-IV criteria by trained clinical psychologists to generate composite scores and percentile ranks against human normative data.

\subsection{Development of the AIQ Benchmark}
\label{subsec:aiq_dev}

To address the limitations of human-normed psychometric instruments and to provide a more sensitive measure of current and future AI capabilities, the AIQ Benchmark was developed. The benchmark is composed of subtests analogous to those in the WAIS-IV, including analogues for Matrix Reasoning, Picture Completion, Figure Weights, and verbal/working memory tasks. All subtests feature novel, procedurally generated item pools designed for scalable difficulty and automated administration.

\subsubsection{AIQ Verbal Reasoning}
\label{subsubsec:aiq_verbal}

Two novel subtests were created to assess AI model performance in the verbal reasoning domain, an area where prior models have achieved ceiling-level performance on existing benchmarks.

\begin{description}[leftmargin=1.5em, itemsep=0.5ex, font=\bfseries]
    \item[AIQ Vocabulary:] This 184-item subtest was developed to extend the difficulty of its WAIS-IV analogue. Word frequency data was used to guide item creation, sampling from the SUBTLEXUS database \citep{brysbaert2009moving}. To increase difficulty, a weighted sampling procedure was implemented to heavily favor low-frequency (rare) over high-frequency (common) English words. The format requires the model to identify the correct definition for a given target word from five multiple-choice options (see Appendix C. for example items).

    \item[AIQ Algebraic Reasoning (text):] This 39-item subtest measures complex deductive and algebraic reasoning. While it lacks a direct WAIS-IV analogue, it was designed as a verbal isomorphic counterpart to the image-based \textit{AIQ Algebraic Reasoning (image)} subtest described below. Both versions utilize an identical mathematical framework: a system of linear equations designed to be "underdetermined" (number of variables > number of equations). This ensures that a definitive solution requires high-level deduction rather than straightforward substitution. Items were developed by translating core algebraic equations into narrative descriptions of balanced scales. In each item, plain-English text describes up to four balanced scales establishing the relative weights of various shapes. Models must then determine the missing value for a final scale by selecting the correct shape combination from five text-based multiple-choice options, which were constrained to ensure a single unique solution.

\end{description}

\subsubsection{AIQ Working Memory}
\label{subsubsec:aiq_working_memory}

Four subtests were developed to assess different facets of working memory capacity and mental manipulation. To evaluate performance, each trial utilizes a four-option multiple-choice format. Distractors are constructed with subtle variations—such as transposed, omitted, or substituted items—to differentiate high-precision recall from approximate pattern matching.

\begin{description}[leftmargin=1.5em, itemsep=0.5ex, font=\bfseries]
    \item[AIQ Numeric Attention:] This 18-item subtest measures basic numeric attention and working memory span capacity. Models are presented with a numerical target sequence and must identify the correct verbatim string from the available multiple choice options.

    \item[AIQ Numeric Manipulation:] This 18-item subtest was designed to measure mental manipulation of numerical information. Models are presented with a numerical target sequence comprised of digits 1-9, and required to select the multiple choice option that correctly depicts the target sequence in reverse order. 

    \item[AIQ Numeric Sequencing:] In this 18-item subtest, models must identify a numerical target sequence reordered into ascending numerical order. This task assesses the ability to simultaneously hold, process, and organize information.

    \item[AIQ Alphanumeric Sequencing:] This 18-item subtest presents a mixed string of letters and numbers. The task requires the model to identify the string correctly reordered with numbers first (in ascending order), followed by letters (in alphabetical order). 

\end{description}

For all four working memory subtests, difficulty was systematically scaled by increasing the sequence length across 18 trials. While the corresponding WAIS-IV subtests peak at maximum lengths of eight items (Digit Span Backward and Letter-Number Sequencing) or nine items (Digit Span Forward and Sequencing), the AIQ versions include sequences ranging from 2 to 1,600 items. This provides a test ceiling significantly beyond human working memory capacity, allowing for the differentiation of performance limits across current and future large language models.

\subsubsection{AIQ Visual-Perceptual Reasoning}
\label{subsubsec:aiq_visual}

Three visual reasoning subtests were developed, for which a completely novel set of stimuli was generated from scratch. This process was guided by the psychometric principles of their WAIS-IV analogues but designed to systematically broaden the range and complexity of the underlying logic.

\begin{description}[leftmargin=1.5em, itemsep=0.5ex, font=\bfseries]
    \item[AIQ Pattern Completion:] For this 45-item, 5-option multiple-choice dataset, the logical principles of the WAIS-IV and Raven’s Progressive Matrices were analyzed. While retaining foundational structures (e.g., 2x2 progressions), a new stimulus set was procedurally generated to include more complex logical rules, such as mathematical progressions (e.g., doubling functions, fibonacci sequence), multi-step rotational/color patterns, and complex figure-ground relationships (see Appendix A).

    \item[AIQ Algebraic Reasoning (image):] This 39-item test was developed based on the quantitative reasoning principles of the WAIS-IV version. The novel item set was created with two key differences: (1) the maximum number of scales per item was increased from three to four, and (2) items included fractional shapes, requiring calculations with non-integer values.

    \item[AIQ Anomaly Detection:] For this 49-item subtest, a new set of stimuli was created via a two-stage process. First, complete images (both line drawings and photo-realistic scenes) were generated using Imagen 3 \citep{baldridge2024imagen}, a latent diffusion text-to-image model. Second, each image was post-processed in Adobe\textsuperscript\textregistered Photoshop \citep{Photoshop26.3.0} to digitally remove an essential visual element. This methodology resulted in a novel item set with a larger size and more subtle, contextually integrated missing elements than the WAIS-IV original.

\end{description}

\subsubsection{Test size and item validation}
\label{subsubsec:test_size}

To evaluate the performance stability and effectiveness of difficulty scaling across each of the nine individual AIQ Benchmark subtests, we analyzed bootstrapped accuracy distributions across sequential item subsets. For each of the nine subtests, we constructed sampling distributions for three cumulative test lengths: Small (the first 1/3 of items), Medium (the first 2/3), and Full (the complete item set). Distributions were generated by bootstrapping the mean accuracy across models using 2,000 resamples with replacement from the item-level response data (\autoref{fig:bootstrapped_dist}). This analysis demonstrated that as the cumulative test set grew and incorporated increasingly difficult items, the distributions transitioned from high-accuracy ceiling effects to approximately normal distributions. This convergence indicates that the AIQBench subtests successfully elicit performance variance consistent with traditional psychometric paradigms.

Next, we examined the slope and variance of aggregated performance in relation to progressively difficult test questions within each AIQ subtest. All subtests demonstrated significant slopes across item difficulty in relation to performance (\autoref{fig:aiq_fig4}), with the exception of the AIQ Vocabulary subtest, which demonstrated a flat slope across questions. This indicates that models found AIQ Vocabulary test questions roughly equally easy, even as the frequency of occurrence in the English language decreased across test items.

\subsection{Data and Implementation}
\label{subsec:data_implementation}

The AIQ Benchmark was implemented on top of Google Cloud APIs. This integration allowed for standardized administration of tests to a diverse range of generative AI models, leveraging existing infrastructure for model inference and output capture. Data analysis and visualization were primarily performed using Python. Custom scripts were developed to process model responses, calculate accuracy scores, and generate various descriptive statistics. Libraries such as Pandas were utilized for data manipulation and Matplotlib and Seaborn for creating visualizations, including the histograms and performance graphs presented in the results.

As described in the Statistical Analyses section, Generalized Linear Mixed Models (GLMMs) were run in R using the \texttt{lme4} package. The \texttt{lme4} package provided the necessary functionality to specify random intercepts for \texttt{AIQBench\_Test} and \texttt{Item\_ID}, thus appropriately controlling for variability in test and item difficulty. Post-hoc analyses and significance testing were also conducted within the R environment.

\subsection{Statistical Analyses}
\label{subsec:statistical_analyses}

To formally assess performance differences across Gemini model versions on the AIQ Benchmark, a Generalized Linear Mixed Model (GLMM) with a binomial error distribution (logit link) was specified. The model predicted the log-odds of a correct item response based on the model version, while accounting for variability across tests and items. The GLMM specification included time\_code as a fixed effect, an ordinal factor representing the six Gemini models (1.5 Flash, 1.5 Pro, 2.5 Flash, 2.5 Pro, 3.0 Flash, 3.1 Pro), with Flash 1.5 serving as the reference level. To account for non-independence in the data, the model also included random intercepts for \texttt{AIQBench\_Test}, allowing baseline difficulty to vary across the nine different subtests, and for \texttt{Item\_ID}, allowing difficulty to vary across the 428 unique items within those tests comprising the AIQ Benchmark dataset.
\section{Results}
\label{sec:results}

\subsection{Phase 1: Cross-Model Comparison on Human-Normed WAIS-IV Benchmarks}
\label{subsec:phase1}

Initial assessments using adapted WAIS-IV subtests were conducted across a range of leading generative AI models (\autoref{tab:wais-performance}). While there was some variability in overall performance---with newer models like Claude 3.5 Sonnet and GPT-4o showing stronger VCI and PRI scores---a highly consistent pattern of profound cognitive gaps emerged across all models. Performance on the Verbal Comprehension Index (VCI) and Working Memory Index (WMI) was exceptionally high, frequently placing models at or above the \nth{99} percentile. For example, Claude 3.5 Sonnet achieved a VCI of 145 (99.\nth{9} percentile), while several models scored at the ceiling (150) on the WMI. This indicated exceptional capabilities in retrieving stored verbal knowledge (Vocabulary, Comprehension, Information) and manipulating symbolic information (Digit Span, Arithmetic).

In stark contrast, performance on the Perceptual Reasoning Index (PRI) was consistently at or near floor levels. With the notable exception of Claude 3.5 Sonnet (PRI=81, \nth{10} percentile), all other models scored below the 1st percentile on the PRI (\autoref{tab:wais-performance}). Subtests like Matrix Reasoning, Visual Puzzles, and Picture Completion yielded scaled scores below the \nth{1} percentile for nearly all models, indicating a profound and general deficit in non-verbal, fluid, and organizational reasoning. These cross-model results confirmed that even as some capabilities advance, the architectural divide between linguistic/symbolic processing and perceptual reasoning remains a fundamental limitation.

\subsection{Phase 2: Tracking Generational Improvement with the AIQ Benchmark}
\label{subsec:phase2}

Performance on the AIQ Benchmark was analyzed across several generations of Gemini models (1.5 Flash, 1.5 Pro, 2.5 Flash, 2.5 Pro, 3.0 Flash, 3.1 Pro), revealing significant but uneven cognitive development (\autoref{fig:gemini_trajectories}). Detailed accuracy scores and confidence intervals for each AIQ subtest across Gemini models are presented in \autoref{tab:accuracy-scores}. Performance on the AIQ Benchmark was also evaluated across several generations of OpenAI models (o1-mini, o1, o3-mini, o3, gpt-5.4-mini, and gpt-5.4), revealing a similar pattern of uneven cognitive development in the visual reasoning domain (see: \autoref{fig:aiq_fig3} for a comparison between Gemini and OpenAI model family performance trends).

To formally assess performance differences across Gemini model versions on the AIQ Benchmark, a Generalized Linear Mixed Model (GLMM) with a binomial error distribution (logit link) was specified. The model predicted the log-odds of a correct item response based on the model version, while accounting for variability across tests and items (see \autoref{fig:GLMM} for a comparison of the GLMM estimated trajectory compared to the raw test score performance).

The GLMM specification included time\_code as a fixed effect, an ordinal factor representing the six Gemini models (1.5 Flash, 1.5 Pro, 2.5 Flash, 2.5 Pro, 3.0 Flash, 3.1 Pro), with Gemini 1.5 Flash serving as the reference level. To account for non-independence in the data, the model also included random intercepts for \texttt{AIQBench\_Test}, allowing baseline difficulty to vary across the nine different subtests, and for \texttt{Item\_ID}, allowing difficulty to vary across the 428 unique items within those tests.

GLMM analyses revealed a significant overall effect of model version on accuracy. The random effects variances were substantial for both \texttt{AIQBench\_Test} ($\sigma^2 = 2.56$) and \texttt{Item\_ID} ($\sigma^2 = 3.37$), indicating significant variability in difficulty at both the test and item levels and justifying their inclusion in the model.

GLMM fixed effects results, detailed in \autoref{tab:glmm-results}, revealed a pronounced positive trend in performance with advancing model versions. Relative to the Flash 1.5 baseline, 1.5 Pro demonstrated a significant increase in the odds of a correct response ($\beta = 0.52$, $p = .018$, $OR = 1.68$). Performance gains grew substantially larger with the next generation of models: 2.5 Flash showed a dramatic increase in performance ($\beta = 1.69$, $p < .001$), with the odds of a correct answer being 5.44 times higher than the baseline. This trajectory continued with 2.5 Pro ($\beta = 2.32$, $p < .001$), whose odds of providing a correct response were 10.22 times higher than the reference model. The most significant gains were observed in the most recent iterations; 3.0 Flash ($\beta = 3.47$, $OR = 31.97$) and 3.1 Pro ($\beta = 3.95$, $OR = 51.92$) both showed highly significant improvements ($p < .001$), with 3.1 Pro exhibiting over 50 times the odds of success compared to the Flash 1.5 baseline.

In summary, the GLMM results provide strong evidence that successive versions of the Gemini model yield significant and progressively larger performance gains on the AIQ Benchmark. Even after controlling for inherent test- and item-level difficulty through random effects, the iterative improvements in model architecture and training result in a near-exponential increase in the odds of providing correct responses across the benchmark suite.
\section{Discussion}
\label{sec:discussion}

Our psychometric evaluation reveals a distinct and evolving cognitive profile for large generative models, defined by profound asymmetries. Initial comparisons to human norms established a critical baseline: a stark divide between superlative linguistic/symbolic manipulation and severely impaired perceptual-organizational reasoning. Subsequent analysis with our scalable AIQ benchmark demonstrated that while overall cognitive performance improves dramatically with each model generation, this growth is highly uneven. This uneven evolution suggests that progress is not a uniform march toward general intelligence, but rather may reflect deep-seated architectural biases.

The most telling evidence of this architectural bias is the sharp dissociation between reasoning in linguistic versus visual modalities. Models developed the abstract quantitative reasoning needed for the AIQ Algebraic Reasoning tasks far more effectively when the problem was presented verbally, with performance rapidly approaching mastery. In contrast, the visually analogous version of the same task saw much slower and more constrained growth. This suggests that the models' core strength lies in manipulating structured, language-based symbolic systems, not in a generalized, modality-independent reasoning capacity. This position aligns closely with other leading theorists who highlight new learning under novel conditions and "efficiency of skill acquisition" as core definitional components of intelligent systems \citep{chollet2019measureintelligence}. 

Despite progress in some areas of abstract visual reasoning, such as visual algebraic reasoning, slower performance improvement for tasks like AIQ Anomaly Detection highlight a fundamental architectural bottleneck. This specific failure points to an inability to parse a visual scene for its essential, causally relevant components---a skill that likely relies on a grounded, implicit world model rather than statistical pattern matching. This deficit aligns with findings on LLMs' poor spatial understanding \citep{yamada2024evaluating} and compositional reasoning \citep{alhamoud2025vision}, and it may represent a class of problems that cannot be solved by simply scaling data and compute.

In conclusion, the divide between rapidly scaling verbal abilities and stubbornly weak visuospatial organization suggests that current AGI development approaches may be driving models toward a highly specialized, rather than truly general, form of intelligence \citep{burnell2023measuring}. We speculate that this gap reflects the absence of architectures capable of building integrated, queryable world representations, analogous to the function of the hippocampus in creating cognitive maps \citep{whittington2022how}. Alternatively, the observed performance gaps may also be related to insufficiencies in visual training corpuses impacting downstream multimodal reasoning capabilities. Addressing this imbalance is therefore a critical challenge in the pursuit of truly domain-general artifical intelligence. Future work must move beyond high-level intelligence quotients to assess the more foundational, neurobiologically grounded building blocks of cognition. Probing these core functions will be essential to understanding and overcoming the architectural limitations that currently constrain the path to AGI.

\subsection{Limitations \& Future Directions}
\label{subsec:discussion_scaling}

There are existing limitations to the approach used to benchmark cognitive functions. First, though performance on specific tests may allow researchers to assess improvements in models, assumptions of structural relationships across the human brain and machine learning models are unwarranted. All tasks may differ significantly in the mechanisms that drive cognitive capabilities. 

Next, the current set of AIQ Benchmark tests is unlikely to be sensitive enough to detect change in future generations of models, as we can already observe increasing asymptotic performance across multiple domains. The AIQ Benchmark should be considered a framework that requires evolving datasets that are periodically re-normed against reference models, as was performed in the current study using Gemini 1.5 Flash. To support the goal of tracking model evolution across time, future research should focus on fully synthetic and configurable dataset generation methods to support scalable evaluation benchmarks.

\bibliography{6-main_arxiv}

\appendix
\renewcommand{\thesection}{\Alph{section}}
\renewcommand{\thefigure}{\arabic{figure}}
\renewcommand{\thetable}{\arabic{table}}

\titleformat{\section}
  {\normalfont\Large\bfseries} 
  {Appendix \thesection:}      
  {0.5em}                        
  {#1}

\setcounter{figure}{0}
\setcounter{table}{0}

\vfill 
\section{Figures} 
\label{sec:appendix_figures}

\clearpage
\newgeometry{
    top=1.6cm,    
    bottom=1.6cm, 
    left=1.5cm, 
    right=1.5cm,
}

\setlength{\headwidth}{\textwidth}
\fancyhfoffset[L,R]{0pt}

\begin{landscape} 

    \centering 
    \vspace*{-1.5cm} 
    \begin{adjustwidth}{0cm}{} 
        \centering
        \begin{tikzpicture}
            \node[anchor=north west, inner sep=0] (img) at (0,0) {
                \includegraphics[width=1.47\textwidth]{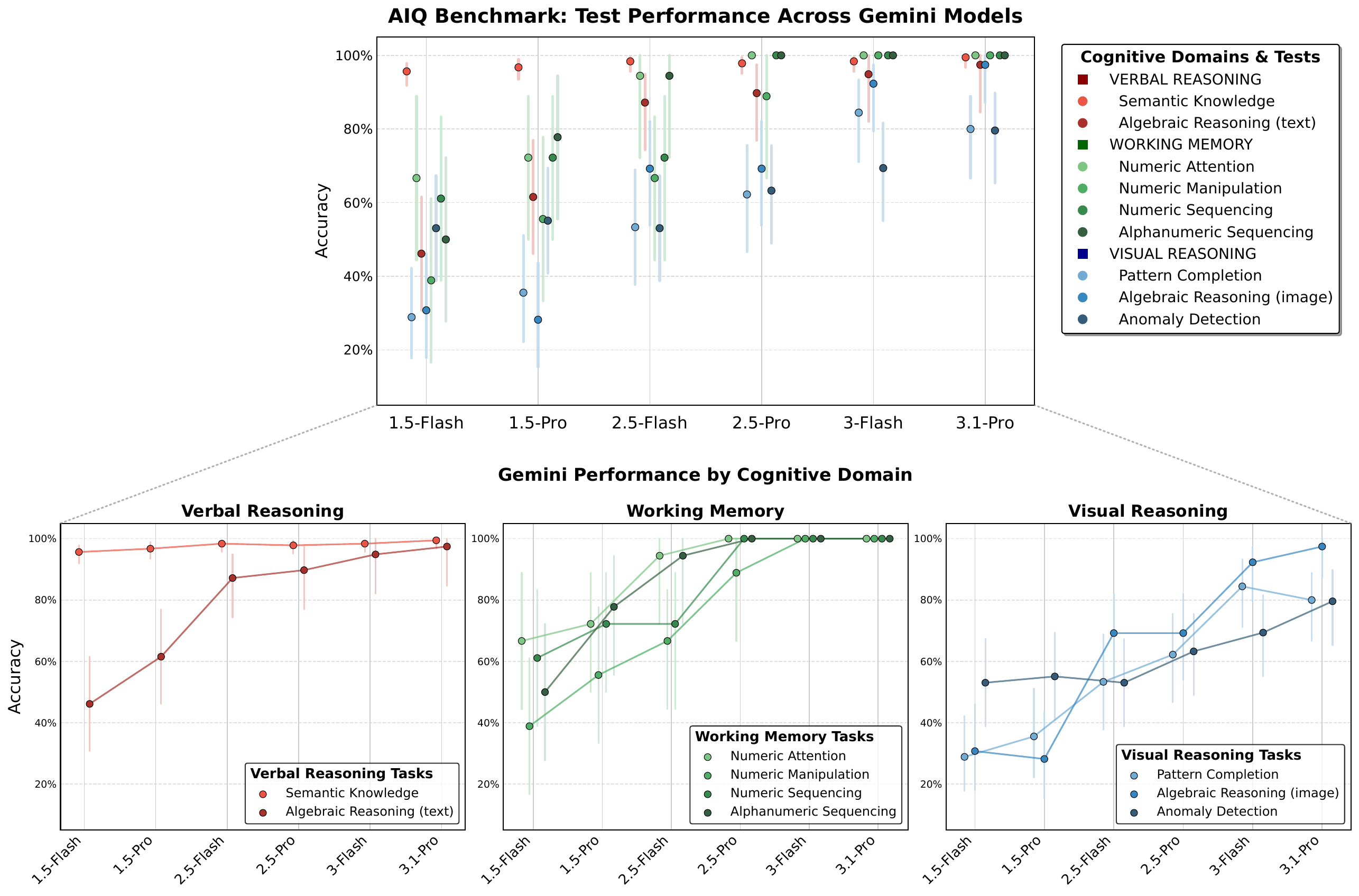}
            };
        \end{tikzpicture}
        \vspace{-.4cm}
        \captionof{figure}{Trajectories of AIQ Benchmark test performances across Gemini model revolutions. Models are presented from left to right in chronological order by release date. Error bars represent 95\% CIs (bootstrapped).}
        \label{fig:gemini_trajectories}
    \end{adjustwidth}

    \clearpage 
    
    \begin{figure}[htbp]
    \begin{adjustwidth}{-.3cm}{} 
        \includegraphics[width=1.5\textwidth]{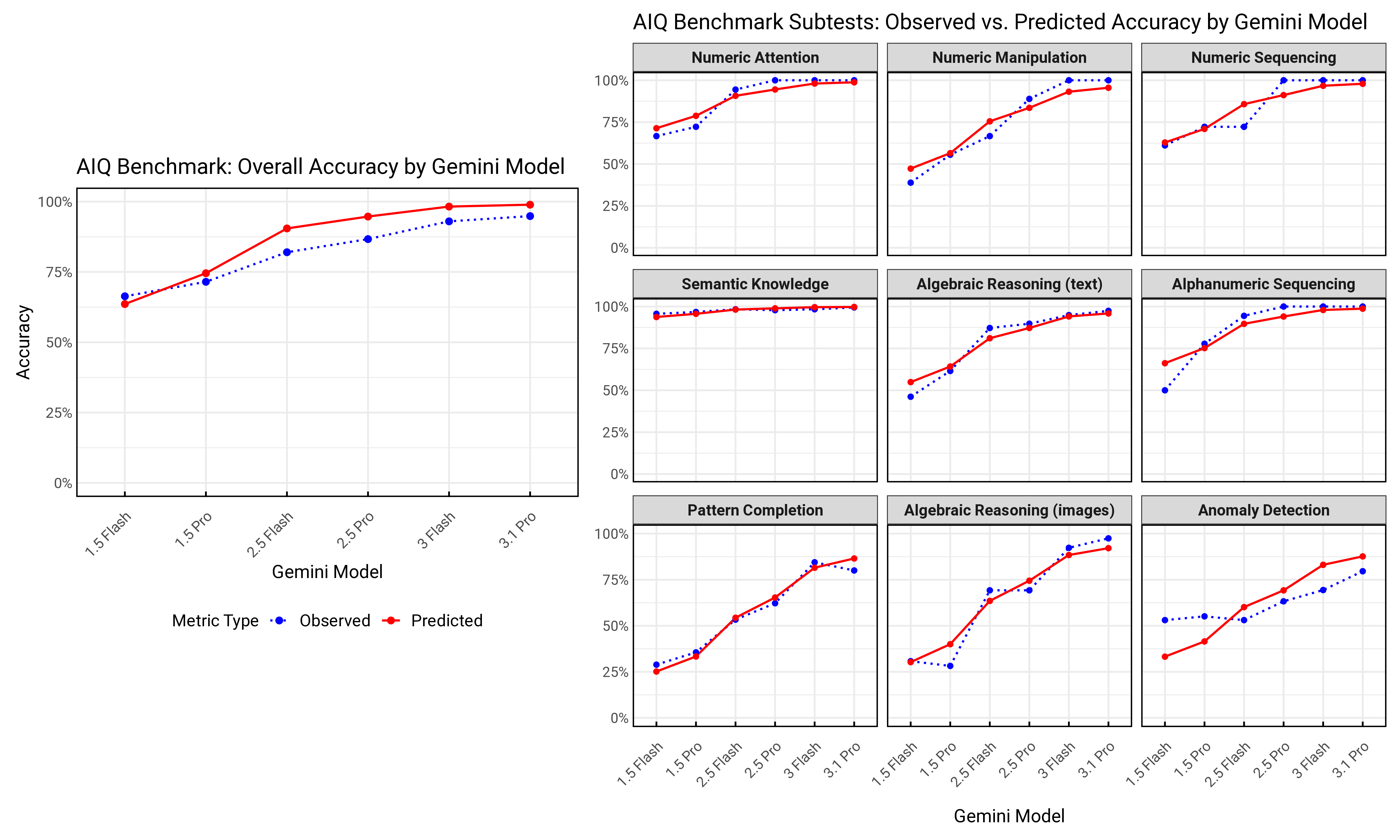}
    \end{adjustwidth}
        \caption{Generalized Linear Mixed Model Results: AIQ Benchmark performance (observed vs. predicted) for the overall model (left). Right-hand pane depicts observed vs. predicted accuracy for the nine AIQ Benchmark subtests (random effect). Gemini models are presented in chronological order by release date.}
        \label{fig:GLMM} 
    \end{figure}
    
    \clearpage 
    
    \begin{figure}[htbp]
    \begin{adjustwidth}{-0.3cm}{} 
        \includegraphics[width=1.5\textwidth]{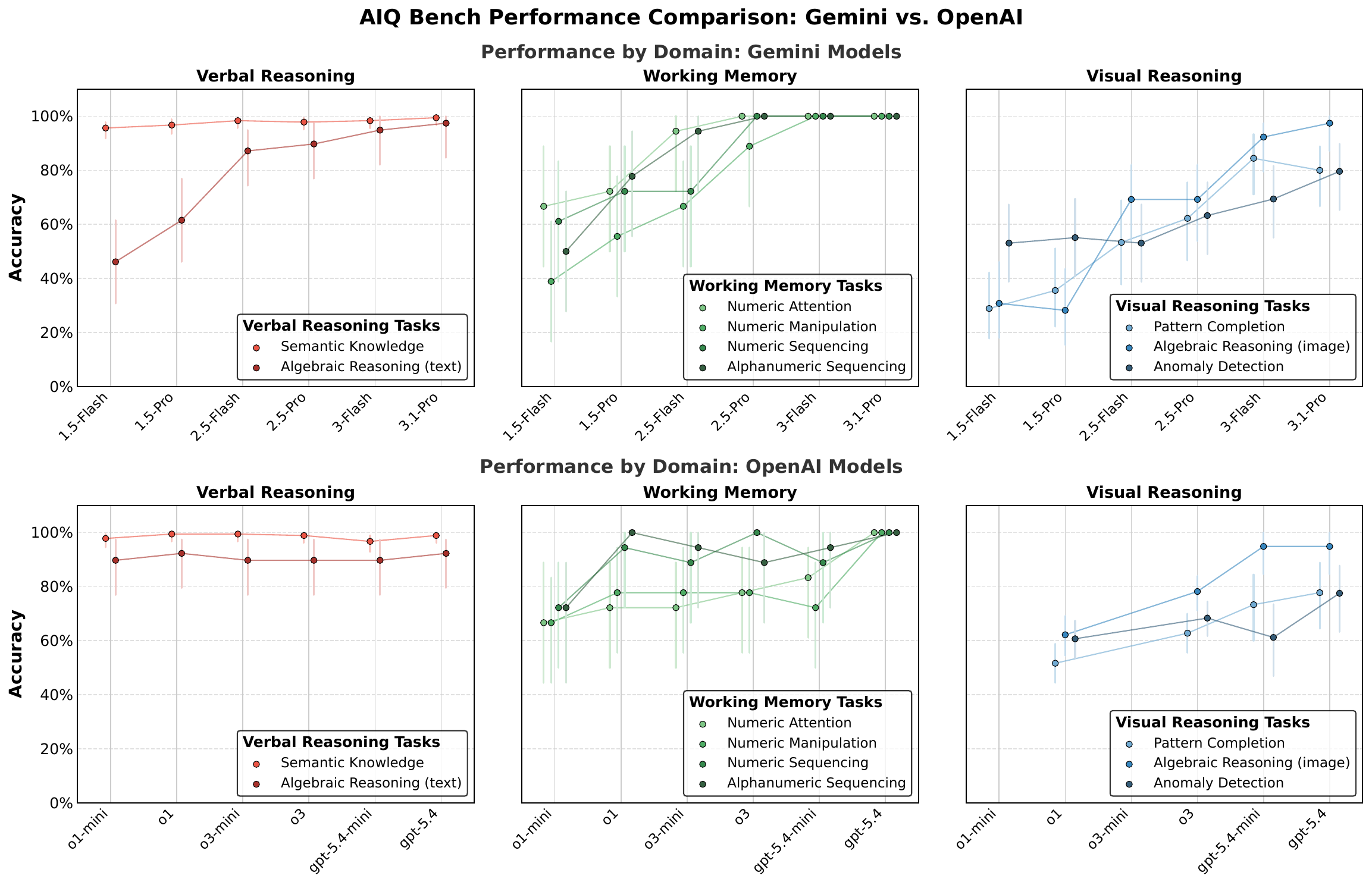}
    \end{adjustwidth}
        \vspace*{-.5cm}
        \caption{Gemini and OpenAI model performance comparisons. Models are presented from left to right in chronological by release date. Error bars represent 95\% CIs (bootstrapped). Note: Visual reasoning scores are unavailable for OpenAI o1-mini and o3-mini due to lack of multimodal capabilities for these models.}
        \label{fig:aiq_fig3} 
    \end{figure}

    \clearpage

    \newpage
    \begin{figure}[htbp]
    \vspace{-1cm}
    \begin{adjustwidth}{}{} 
        \centering
        \includegraphics[width=1.1\textwidth]{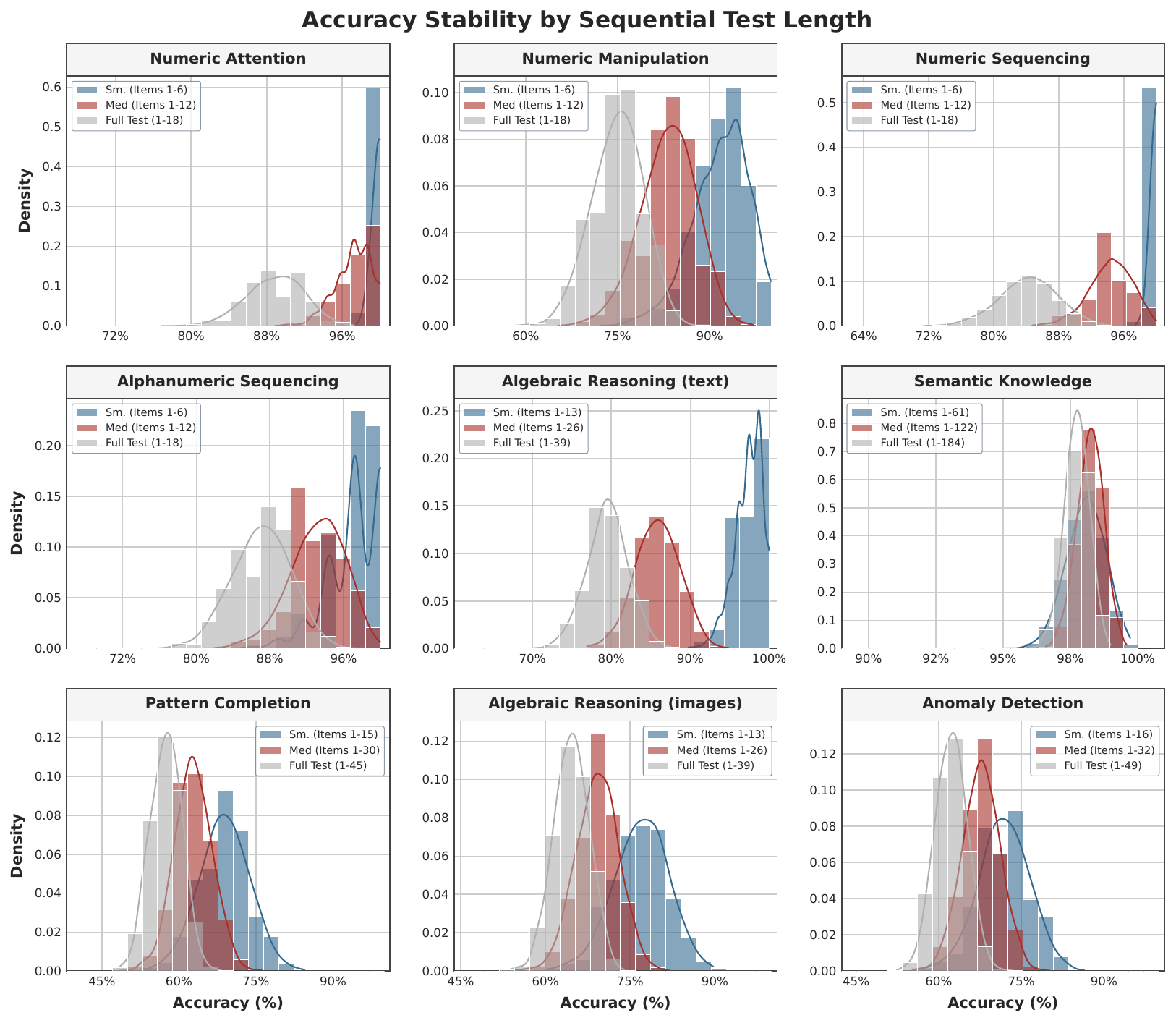}
    \end{adjustwidth}
        \vspace{-.4cm}
        \caption{{Impact of Difficulty Grading on Accuracy Distributions.} Histograms represent bootstrapped sampling distributions for cumulative item sets: Small (first 1/3), Medium (first 2/3), and Full (complete subtest). The distinct separation and trend toward progressive widening of distributions illustrates the success of the difficulty gradient; initial items show higher, low-variance accuracy, while the inclusion of increasingly complex items shifts the mean downward and expands performance variance (with the exception of Semantic Knowledge, where all item set sizes demonstrate similar distributions due to ceiling level performance).}
        \label{fig:bootstrapped_dist} 
    \end{figure}

    \begin{figure}[htbp]
    \vspace*{-1cm} 
    \begin{adjustwidth}{}{} 
        \centering
        \includegraphics[width=1.1\textwidth]{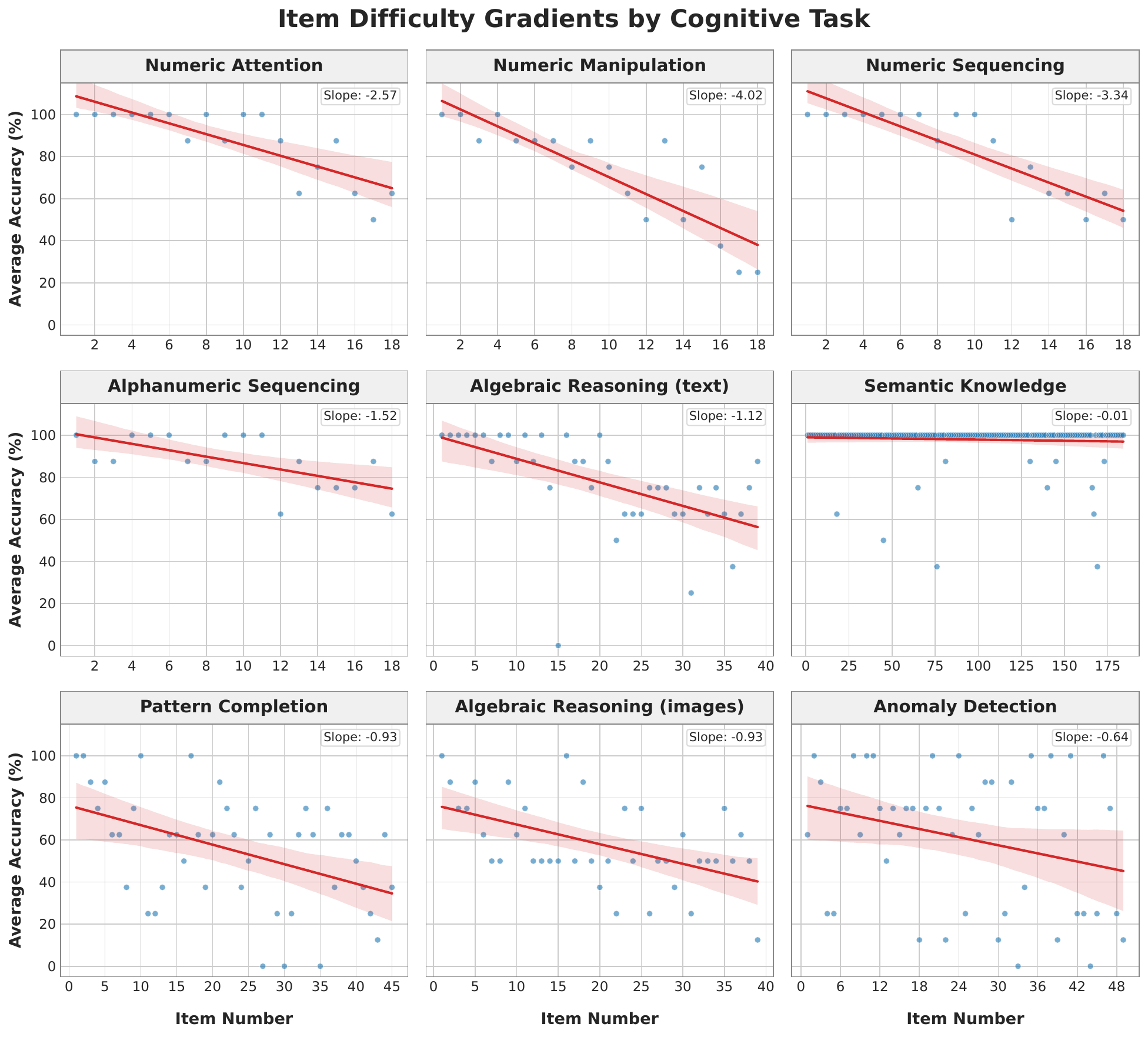}
    \end{adjustwidth}
        \caption{Slope and variance of aggregated performance in relation to progressively
difficult test questions within each AIQ subtest.}
        \label{fig:aiq_fig4} 
    \end{figure}

\clearpage



\section{\raggedright Tables}
\label{sec:appendix_tables}

\begin{table}[htbp]
\caption{Comparative WAIS-IV Performance of Selected Multimodal Generative AI Models.}
\label{tab:wais-performance}
\renewcommand{\arraystretch}{1.3}
\setlength{\tabcolsep}{10pt} 
\begin{tabular}{@{} l cccccc @{}}
\toprule
& \multicolumn{2}{c}{\textbf{OpenAI}} & \multicolumn{2}{c}{\textbf{Google}} & \multicolumn{2}{c}{\textbf{Anthropic}} \\
\cmidrule(lr){2-3} \cmidrule(lr){4-5} \cmidrule(lr){6-7}
\textbf{Cognitive Domain \& Subtest} & \textbf{GPT-4 Turbo} & \textbf{GPT-4o} & \textbf{Gemini 1.5 Flash} & \textbf{Gemini 1.5 Pro} & \textbf{Claude 3 Opus} & \textbf{Claude 3.5 Sonnet} \\
& \textit{Score (\%ile)} & \textit{Score (\%ile)} & \textit{Score (\%ile)} & \textit{Score (\%ile)} & \textit{Score (\%ile)} & \textit{Score (\%ile)} \\
\midrule

\textbf{Verbal Comprehension (VCI)} & \textbf{136 (99)} & \textbf{141 (99.7)} & \textbf{114 (82)} & \textbf{141 (99.7)} & \textbf{138 (99)} & \textbf{145 (99.9)} \\
\quad Similarities & 14 (91) & 15 (95) & 14 (91) & 14 (91) & 13 (84) & 14 (91) \\
\quad Vocabulary & 15 (95) & 16 (98) & 6 (9) & 17 (99) & 17 (99) & 19 ($>$99.9) \\
\quad Information & 19 ($>$99.9) & 19 ($>$99.9) & 19 ($>$99.9) & 19 ($>$99.9) & 19 ($>$99.9) & 19 ($>$99.9) \\
\quad Comprehension & 18 (99.6) & 19 ($>$99.9) & 19 ($>$99.9) & 19 ($>$99.9) & 17 (99) & 19 ($>$99.9) \\
\addlinespace

\textbf{Working Memory Index (WMI)} & \textbf{150 ($>$99.9)} & \textbf{150 ($>$99.9)} & \textbf{139 (99.5)} & \textbf{145 (99.9)} & \textbf{150 ($>$99.9)} & \textbf{145 (99.9)} \\
\quad Digit Span & 19 ($>$99.9) & 19 ($>$99.9) & 19 ($>$99.9) & 19 ($>$99.9) & 19 ($>$99.9) & 19 ($>$99.9) \\
\quad Arithmetic & 19 ($>$99.9) & 19 ($>$99.9) & 15 (95) & 17 (99) & 19 ($>$99.9) & 17 (99) \\
\quad Letter-Number Seq. & 19 ($>$99.9) & 19 ($>$99.9) & 16 (98) & 19 ($>$99.9) & 14 (91) & 16 (98) \\
\addlinespace

\textbf{Perceptual Reasoning (PRI)} & \textbf{50 ($<$0.1)} & \textbf{58 (0.3)} & \textbf{50 ($<$0.1)} & \textbf{50 ($<$0.1)} & \textbf{52 ($<$0.1)} & \textbf{81 (10)} \\
\quad Matrix Reasoning & 1 (0.1) & 1 (0.1) & 1 (0.1) & 1 (0.1) & 1 (0.1) & 8 (25) \\
\quad Visual Puzzles & 1 (0.1) & 2 (0.4) & 1 (0.1) & 1 (0.1) & 0 ($<$0.1) & 2 (0.4) \\
\quad Figure Weights & 2 (0.4) & 6 (9) & 2 (0.4) & 4 (2) & 1 (0.1) & 10 (50) \\
\quad Picture Completion & 1 (0.1) & 1 (0.1) & 1 (0.1) & 1 (0.1) & 1 (0.1) & 1 (0.1) \\

\bottomrule
\end{tabular}

\vspace{2mm}
\begin{minipage}{1.4\textwidth}
\textit{Note:} Index scores (VCI, WMI, PRI) use Standard Score format ($M=100$, $SD=15$), while individual subtests use Scaled Scores ($M=10$, $SD=3$).
\end{minipage}
\end{table}

\clearpage

\begin{table}[htbp]
\caption{Accuracy Scores (\%) on AIQ Subtests Across Six Gemini Model Generations.}
\label{tab:accuracy-scores}
\renewcommand{\arraystretch}{1.3}
\setlength{\tabcolsep}{4pt} 
\begin{tabular}{l ccccc ccccc cc}
\toprule
& \multicolumn{2}{c}{\textbf{1.5 Flash}} & \multicolumn{2}{c}{\textbf{1.5 Pro}} & \multicolumn{2}{c}{\textbf{2.5 Flash}} & \multicolumn{2}{c}{\textbf{2.5 Pro}} & \multicolumn{2}{c}{\textbf{3.0 Flash}} & \multicolumn{2}{c}{\textbf{3.1 Pro}} \\
\cmidrule(lr){2-3} \cmidrule(lr){4-5} \cmidrule(lr){6-7} \cmidrule(lr){8-9} \cmidrule(lr){10-11} \cmidrule(lr){12-13}
AIQ Subtests & Acc. & 95\% CI & Acc. & 95\% CI & Acc. & 95\% CI & Acc. & 95\% CI & Acc. & 95\% CI & Acc. & 95\% CI \\
\midrule
\multicolumn{13}{l}{\textbf{Verbal Reasoning}} \\
\quad Vocabulary & 95.7\% & [91.8, 97.8] & 96.7\% & [93.5, 98.9] & 98.4\% & [95.7, 99.5] & 97.8\% & [95.1, 99.5] & 98.37\% & [95.6, 99.4] & 99.46\% & [96.7, 100] \\
\quad Algebraic (text) & 46.2\% & [30.8, 61.5] & 61.5\% & [46.2, 76.9] & 87.2\% & [74.4, 94.9] & 89.7\% & [76.9, 97.4] & 94.87\% & [82.0, 100] & 97.44\% & [84.6, 100] \\
\addlinespace

\multicolumn{13}{l}{\textbf{Working Memory}} \\
\quad Numeric Attention & 66.7\% & [44.4, 88.9] & 72.2\% & [50.0, 88.9] & 94.4\% & [72.2, 100] & 100.0\% & [100, 100] & 100.0\% & [100, 100] & 100.0\% & [100, 100] \\
\quad Numeric Manip. & 38.9\% & [16.7, 61.1] & 55.6\% & [33.3, 77.8] & 66.7\% & [44.4, 83.3] & 88.9\% & [66.7, 100] & 100.0\% & [100, 100] & 100.0\% & [100, 100] \\
\quad Numeric Seq. & 61.1\% & [38.9, 83.3] & 72.2\% & [50.0, 88.9] & 72.2\% & [44.4, 88.9] & 100.0\% & [100, 100] & 100.0\% & [100, 100] & 100.0\% & [100, 100] \\
\quad Alphanum. Seq. & 50.0\% & [27.8, 72.2] & 77.8\% & [55.6, 94.4] & 94.4\% & [72.2, 100] & 100.0\% & [100, 100] & 100.0\% & [100, 100] & 100.0\% & [100, 100] \\
\addlinespace

\multicolumn{13}{l}{\textbf{Visual Reasoning}} \\
\quad Pattern Comp. & 28.9\% & [17.8, 42.2] & 35.6\% & [22.2, 51.1] & 53.3\% & [37.8, 68.9] & 62.2\% & [46.7, 75.6] & 84.44\% & [71.1, 93.3] & 80.00\% & [66.6, 88.8] \\
\quad Algebraic (image) & 30.8\% & [17.9, 46.2] & 28.2\% & [15.4, 43.6] & 69.2\% & [53.8, 82.1] & 69.2\% & [53.8, 82.1] & 92.31\% & [79.4, 97.4] & 97.44\% & [87.1, 100] \\
\quad Anomaly Det. & 53.1\% & [38.8, 67.3] & 55.1\% & [40.8, 69.4] & 53.1\% & [38.8, 67.3] & 63.3\% & [49.0, 75.5] & 69.39\% & [55.1, 81.6] & 79.59\% & [65.3, 89.8] \\
\bottomrule
\end{tabular}

\vspace{2mm}
\textit{Note:} All values are percentages representing mean accuracy scores. Bracketed values indicate the 95\% confidence interval [Lower Bound, Upper Bound]. 
\end{table}

\clearpage

\begin{table}[htbp]
\begin{minipage}{1.35\textwidth}
\caption{Generalized Linear Mixed Effects Model Results. Model specification included fixed effect for time\_code (i.e., Gemini model version, ordered by release date), and random effects for \texttt{AIQBench\_Test} and \texttt{Item\_ID}.}
\label{tab:glmm-results}
\end{minipage}

\renewcommand{\arraystretch}{1.3}
\setlength{\tabcolsep}{14pt} 

\begin{tabular}{>{\raggedright\arraybackslash}p{2.5in} cccccc}
\toprule
Predictor & Estimate ($\beta$) & Std. Error & $z$-value & $p$-value & Odds Ratio & 95\% CI \\
\midrule
Intercept & 0.56 & 0.57 & 0.98 & .33 & -- & -- \\
\addlinespace
\multicolumn{7}{l}{\textbf{Model Version} (Reference: Flash 1.5)} \\
\quad 1.5 Pro (time\_code1) & 0.52 & 0.22 & 2.37 & .018* & 1.68 & (1.09, 2.57) \\
\quad 2.5 Flash (time\_code2) & 1.69 & 0.24 & 7.06 & <.001*** & 5.44 & (3.40, 8.70) \\
\quad 2.5 Pro (time\_code3) & 2.32 & 0.26 & 8.98 & <.001*** & 10.22 & (6.15, 16.97) \\
\quad 3.0 Flash (time\_code4) & 3.47 & 0.31 & 11.16 & <.001*** & 31.97 & (17.40, 58.75) \\
\quad 3.1 Pro (time\_code5) & 3.95 & 0.34 & 11.57 & <.001*** & 51.92 & (26.59, 101.38) \\
\bottomrule
\end{tabular}

\begin{minipage}{1.35\textwidth}
\vspace{2mm}
\textit{Note:} Odds Ratios (OR) and their 95\% Confidence Intervals (CI) were calculated by exponentiating the coefficients and their corresponding confidence intervals. An OR greater than 1 indicates increased odds of a correct response relative to the reference model (Flash 1.5). Statistical significance thresholds: * $p < .05$, *** $p < .001$.
\end{minipage}
\end{table}

\clearpage

\end{landscape}

\restoregeometry  


\section{AIQ Benchmark Example Items by Cognitive Domain}
\small
\label{sec:appendix_task_examples}

\subsection*{\large I. Verbal Reasoning}

\subsubsection*{\large AIQ Vocabulary (Semantic Knowledge)}

\noindent\textbf{Example A. (easy item):} \textit{“Which one of the following response options (A, B, C, D or E) is the most accurate definition of the following word: BREAK”}

\begin{enumerate}[label=\Alph*., leftmargin=2em] 
    \item To become damaged but still be in working order.
    \item To make something no longer as effective as it was originally, rendering it less useful.
    \item An internal malfunction or failure, often related to user error.
    \item \textbf{To separate into two or more pieces, to fracture or crack, by a process that cannot easily be reversed for reassembly. \textcolor{darkgreen}{(GT)}}
    \item To adjust the continuity or flow of something, without disrupting its progress.
\end{enumerate}

\vspace{0.2cm}

\noindent\textbf{Example B. (difficult item):} \textit{“Which one of the following response options (A, B, C, D or E) is the most accurate definition of the following word: INVESTITURE”}

\begin{enumerate}[label=\Alph*., leftmargin=2em]
    \item The act of establishing authority through military conquest or force.
    \item \textbf{The act of establishing in office or ratifying; the formal bestowal, confirmation, or presentation of rank, office, or rights. \textcolor{darkgreen}{(GT)}}
    \item The act of divesting oneself of possessions or property.
    \item The act of initiating a new and ambitious project or venture.
    \item The act of divesting money in a financial asset.
\end{enumerate}

\vspace{0.3cm}

\subsubsection*{\large AIQ Algebraic Reasoning (text)}

\noindent\textbf{Example A. (easy item):} \textit{“A scale with two trays is perfectly balanced with 3 yellow triangles on one tray, and 3 purple crosses on the other tray. A second scale with two trays has 1 yellow triangle on one of the trays. What should I put on the other tray in order to balance this second scale? Choose from the following response options:”}
\begin{enumerate}[label=\Alph*., leftmargin=2em]
    \item 3 yellow triangles
    \item 1 yellow triangle
    \item \textbf{1 purple cross \textcolor{darkgreen}{(GT)}}
    \item 3 purple crosses
    \item 1 yellow cross
\end{enumerate}

\vspace{0.3cm}

\noindent\textbf{Example B. (difficult item):} \textit{“A scale with two trays is perfectly balanced with 3 pink hearts on one tray, and 1 red rectangle and 1 blue triangle on the other tray. A second scale with two trays is perfectly balanced with 2 red rectangles on one tray, and 2 green squares and 1 pink heart on the other tray. A third scale with two trays is perfectly balanced with 3 green squares on one tray, and 1 pink heart and 1 blue triangle on the other tray. A new fourth scale with two trays has 5 red rectangles on one of the trays. Which of the following response options (A, B, C, D, or E) should I put on the other tray to balance the fourth scale?”}
\begin{enumerate}[label=\Alph*., leftmargin=2em]
    \item \textbf{2 pink hearts and 3 blue triangles \textcolor{darkgreen}{(GT)}}
    \item 5 green squares
    \item 5 green squares and 1 blue triangle
    \item 2 green squares and 1 blue triangle
    \item 4 pink hearts
\end{enumerate}

\clearpage

\subsection*{\Large II. Working Memory}

\subsubsection*{\large AIQ Numeric Attention}

\noindent\textbf{Example A. (easy item):} \textit{“Which one of the answer options below (A, B, C, or D) matches the following string of numbers? 6, 3, 5, 8, 7, 9”}
\begin{enumerate}[label=\Alph*., leftmargin=2em]
    \item \textbf{6, 3, 5, 8, 7, 9 \textcolor{darkgreen}{(GT)}}
    \item 7, 3, 5, 8, 6, 9
    \item 3, 6, 5, 8, 7, 9
    \item 6, 9, 5, 8, 7, 3 
\end{enumerate}

\vspace{0.3cm}

\noindent\textbf{Example B. (difficult item):} \textit{“Which one of the answer options below (A, B, C, or D) matches the following string of numbers? 1, 5, 2, 2, 7, 7, 1, 3, 4, 8, 3, 5, 1, 7, 3, 5, 8, 4, 1, 6, 5, 1, 3, 4, 2, 8, 2, 4, 5, 2, 8, 5, 1, 3, 8, 5, 1, 3, 7, 3, 5, 6, 5, 5, 9, 2, 6, 8, 1, 2, 4, 3, 9, 3, 5, 9, 3, 9, 8, 9, 3, 4, 4, 7, 1, 4, 7, 4, 7, 4, 3, 3, 3, 7, 5, 7, 1, 5, 3, 2, 7, 2, 8, 6, 7, 2, 7, 7, 7, 7, 3, 6, 3, 3, 4, 3, 4, 9, 6, 5, 6, 2, 6, 9, 3, 6, 7, 5, 1, 9, 6, 6, 6, 2, 1, 1, 1, 1, 1, 8, 7, 4, 6, 5, 4, 5, 6, 2, 4, 7, 8, 5, 1, 1, 8, 4, 4, 1, 1, 5, 5, 6, 9, 9, 1, 2, 6, 7, 2, 5, 1, 5, 8, 8, 8, 7, 3, 7, 1, 3, 8, 8, 3, 7, 2, 4, 8, 1, 6, 5, 6, 6, 4, 9, 1, 2, 6, 5, 2, 6, 1, 7, 9, 9, 7, 4, 5, 5, 4, 7, 6, 9, 2, 4, 9, 2, 6, 9, 1, 4, 8, 2, 8, 9, 1, 8, 6, 7, 9, 8, 6, 2, 1, 1, 6, 7, 4, 7, 2, 8, 8, 1, 8, 7, 1, 5, 3, 8, 6, 3, 9, 6, 7, 7, 5, 2, 6, 5, 6, 6, 7, 4, 9, 2, 5, 3, 8, 5, 2, 9, 3, 4, 4, 5, 2, 7, 3, 6, 5, 2, 6, 5, 6, 8, 5, 5, 3, 3, 2, 4, 1, 2, 2, 7, 9, 7, 4, 1, 8, 4, 3, 6, 8, 4, 5, 8, 7, 8, 4, 7, 4, 8, 2, 4, 8, 9, 9, 5, 6, 4, 7, 7, 3, 4, 6, 8, 1, 8, 2, 3, 6, 1, 6, 9, 9, 3, 6, 4, 2, 8, 4, 9, 6, 6, 2, 8, 3, 2, 1, 2, 1, 2, 2, 7, 9, 9, 4, 3, 1, 8, 5, 7, 1, 6, 4, 4, 5, 1, 7, 6, 4, 1, 1, 2, 3, 1, 3, 3, 7, 8, 9, 9, 3, 5, 8, 8, 6, 8, 3, 5, 2, 2, 2, 8, 1, 1, 3, 2, 5, 7, 1, 7, 7, 3, 9, 7, 7, 6, 3, 6, 1, 4, 8, 1, 5, 2, 4, 6, 8, 7”}

{\raggedright \small
\begin{enumerate}[label=\Alph*., leftmargin=*, itemsep=1ex]
    \item 1, 5, 2, 2, 7, 7, 1, 3, 4, 8, 3, 5, 1, 7, 3, 5, 8, 4, 1, 6, 5, 1, 3, 4, 2, 8, 2, 4, 5, 2, 8, 5, 1, 3, 8, 5, 1, 3, 7, 3, 5, 6, 5, 5, 9, 2, 6, 8, 1, 2, 4, 3, 9, 3, 5, 9, 3, 9, 8, 9, 3, 4, 4, 7, 1, 4, 7, 4, 7, 4, 3, 3, 3, 7, 5, 7, 1, 5, 3, 2, 7, 2, 8, 6, 7, 2, 7, 7, 7, 7, 3, 6, 3, 3, 4, 3, 4, 9, 6, 5, 6, 2, 6, 9, 3, 6, 7, 5, 1, 9, 6, 6, 6, 2, 1, 1, 1, 1, 1, 8, 7, 4, 6, 5, 4, 5, 6, 2, 4, 7, 8, 5, 1, 1, 8, 4, 4, 1, 1, 5, 5, 6, 9, 9, 1, 2, 6, 7, 2, 5, 1, 5, 8, 8, 8, 7, 3, 7, 1, 3, 8, 8, 3, 7, 2, 4, 8, 1, 6, 5, 6, 6, 4, 9, 1, 2, 6, 5, 2, 6, 1, 7, 9, 9, 7, 4, 5, 5, 4, 7, 6, 9, 2, 4, 9, 2, 6, 9, 1, 4, 8, 2, 8, 9, 1, 8, 6, 7, 9, 8, 6, 2, 1, 1, 6, 7, 4, 7, 2, 8, 8, 1, 8, 7, 1, 5, 3, 8, 6, 3, 9, 6, 7, 7, 5, 2, 6, 5, 6, 6, 7, 4, 9, 2, 5, 3, 8, 5, 2, 9, 3, 4, 4, 5, 2, 7, 3, 6, 5, 2, 6, 5, 6, 8, 5, 5, 3, 3, 2, 4, 1, 2, 2, 7, 9, 7, 4, 1, 8, 4, 3, 6, 8, 4, 5, 8, 7, 8, 4, 7, 4, 8, 2, 4, 8, 9, 9, 5, 6, 4, 7, 7, 3, 4, 6, 8, 1, 8, 2, 3, 6, 1, 6, 9, 9, 3, 6, 4, 2, 8, 4, 9, 6, 6, 2, 8, 3, 2, 1, 2, 1, 2, 2, 7, 9, 9, 4, 3, 1, 8, 5, 7, 1, 6, 4, 4, 5, 1, 7, 6, 4, 1, 1, 2, 3, 1, 3, 3, 7, 8, 7, 9, 3, 5, 8, 8, 6, 8, 3, 5, 2, 2, 2, 8, 1, 1, 3, 2, 5, 7, 1, 7, 9, 3, 9, 7, 7, 6, 3, 6, 1, 4, 8, 1, 5, 2, 4, 6, 8, 7
    \item \textbf{1, 5, 2, 2, 7, 7, 1, 3, 4, 8, 3, 5, 1, 7, 3, 5, 8, 4, 1, 6, 5, 1, 3, 4, 2, 8, 2, 4, 5, 2, 8, 5, 1, 3, 8, 5, 1, 3, 7, 3, 5, 6, 5, 5, 9, 2, 6, 8, 1, 2, 4, 3, 9, 3, 5, 9, 3, 9, 8, 9, 3, 4, 4, 7, 1, 4, 7, 4, 7, 4, 3, 3, 3, 7, 5, 7, 1, 5, 3, 2, 7, 2, 8, 6, 7, 2, 7, 7, 7, 7, 3, 6, 3, 3, 4, 3, 4, 9, 6, 5, 6, 2, 6, 9, 3, 6, 7, 5, 1, 9, 6, 6, 6, 2, 1, 1, 1, 1, 1, 8, 7, 4, 6, 5, 4, 5, 6, 2, 4, 7, 8, 5, 1, 1, 8, 4, 4, 1, 1, 5, 5, 6, 9, 9, 1, 2, 6, 7, 2, 5, 1, 5, 8, 8, 8, 7, 3, 7, 1, 3, 8, 8, 3, 7, 2, 4, 8, 1, 6, 5, 6, 6, 4, 9, 1, 2, 6, 5, 2, 6, 1, 7, 9, 9, 7, 4, 5, 5, 4, 7, 6, 9, 2, 4, 9, 2, 6, 9, 1, 4, 8, 2, 8, 9, 1, 8, 6, 7, 9, 8, 6, 2, 1, 1, 6, 7, 4, 7, 2, 8, 8, 1, 8, 7, 1, 5, 3, 8, 6, 3, 9, 6, 7, 7, 5, 2, 6, 5, 6, 6, 7, 4, 9, 2, 5, 3, 8, 5, 2, 9, 3, 4, 4, 5, 2, 7, 3, 6, 5, 2, 6, 5, 6, 8, 5, 5, 3, 3, 2, 4, 1, 2, 2, 7, 9, 7, 4, 1, 8, 4, 3, 6, 8, 4, 5, 8, 7, 8, 4, 7, 4, 8, 2, 4, 8, 9, 9, 5, 6, 4, 7, 7, 3, 4, 6, 8, 1, 8, 2, 3, 6, 1, 6, 9, 9, 3, 6, 4, 2, 8, 4, 9, 6, 6, 2, 8, 3, 2, 1, 2, 1, 2, 2, 7, 9, 9, 4, 3, 1, 8, 5, 7, 1, 6, 4, 4, 5, 1, 7, 6, 4, 1, 1, 2, 3, 1, 3, 3, 7, 8, 9, 9, 3, 5, 8, 8, 6, 8, 3, 5, 2, 2, 2, 8, 1, 1, 3, 2, 5, 7, 1, 7, 7, 3, 9, 7, 7, 6, 3, 6, 1, 4, 8, 1, 5, 2, 4, 6, 8, 7 \textcolor{darkgreen}{(GT)}}
    \item 1, 5, 2, 2, 7, 7, 1, 3, 4, 8, 3, 5, 1, 7, 3, 5, 8, 4, 1, 6, 5, 1, 3, 4, 2, 8, 2, 4, 5, 2, 8, 5, 1, 3, 8, 5, 1, 3, 7, 3, 5, 6, 5, 5, 9, 2, 6, 8, 1, 2, 4, 3, 9, 3, 5, 9, 3, 9, 8, 9, 3, 4, 4, 7, 1, 4, 7, 4, 7, 4, 3, 3, 3, 7, 5, 7, 1, 5, 3, 2, 7, 2, 8, 6, 7, 2, 7, 7, 7, 7, 3, 6, 3, 3, 4, 3, 4, 9, 6, 5, 6, 2, 6, 9, 3, 6, 7, 5, 1, 9, 6, 6, 6, 2, 1, 1, 1, 1, 5, 8, 7, 4, 6, 5, 4, 5, 6, 2, 4, 7, 8, 5, 1, 1, 8, 4, 4, 1, 1, 5, 5, 6, 9, 9, 1, 2, 6, 7, 2, 5, 1, 5, 8, 8, 8, 7, 3, 7, 1, 3, 8, 8, 3, 7, 2, 4, 8, 1, 6, 5, 6, 6, 4, 9, 1, 2, 6, 1, 2, 6, 1, 7, 9, 9, 7, 4, 5, 5, 4, 7, 6, 9, 2, 4, 9, 2, 6, 9, 1, 4, 8, 2, 8, 9, 1, 8, 6, 7, 9, 8, 6, 2, 1, 1, 6, 7, 4, 7, 2, 8, 8, 1, 8, 7, 1, 5, 3, 8, 6, 3, 9, 6, 7, 7, 5, 2, 6, 5, 6, 6, 7, 4, 9, 2, 5, 3, 8, 5, 2, 9, 3, 4, 4, 5, 2, 7, 3, 6, 5, 2, 6, 5, 6, 8, 5, 5, 3, 3, 2, 4, 1, 2, 2, 7, 9, 7, 4, 1, 8, 4, 3, 6, 8, 4, 5, 8, 7, 8, 4, 7, 4, 8, 2, 4, 8, 9, 9, 5, 6, 4, 7, 7, 3, 4, 6, 8, 1, 8, 2, 3, 6, 1, 6, 9, 9, 3, 6, 4, 2, 8, 4, 9, 6, 6, 2, 8, 3, 2, 1, 2, 1, 2, 2, 7, 9, 9, 4, 3, 1, 8, 5, 7, 1, 6, 4, 4, 5, 1, 7, 6, 4, 1, 1, 2, 3, 1, 3, 3, 7, 8, 9, 9, 3, 5, 8, 8, 6, 8, 3, 5, 2, 2, 2, 8, 1, 1, 3, 2, 5, 7, 1, 7, 7, 3, 9, 7, 7, 6, 3, 6, 1, 4, 8, 1, 5, 2, 4, 6, 8, 7
    \item 1, 5, 2, 2, 7, 7, 1, 3, 4, 8, 3, 5, 1, 7, 3, 5, 8, 4, 1, 6, 5, 1, 3, 4, 2, 8, 2, 4, 5, 2, 8, 5, 1, 3, 8, 5, 1, 3, 7, 3, 5, 6, 5, 5, 9, 2, 6, 8, 1, 2, 4, 3, 9, 3, 5, 9, 3, 9, 8, 9, 3, 4, 4, 7, 1, 4, 7, 4, 7, 4, 3, 3, 3, 7, 5, 7, 1, 5, 3, 2, 7, 2, 8, 6, 7, 2, 7, 7, 7, 7, 3, 6, 3, 3, 4, 3, 4, 9, 6, 5, 6, 2, 6, 9, 3, 6, 7, 5, 1, 9, 6, 6, 6, 2, 1, 1, 1, 1, 1, 8, 7, 4, 6, 5, 4, 5, 6, 2, 4, 7, 8, 5, 1, 1, 8, 4, 4, 1, 1, 5, 5, 9, 9, 9, 1, 2, 6, 7, 2, 5, 1, 5, 8, 8, 8, 7, 3, 7, 1, 3, 8, 8, 3, 7, 2, 4, 8, 1, 6, 5, 6, 6, 4, 9, 1, 2, 6, 5, 2, 6, 1, 7, 9, 9, 7, 4, 5, 5, 4, 7, 6, 9, 2, 4, 9, 2, 6, 9, 1, 4, 8, 2, 8, 9, 1, 8, 6, 7, 9, 8, 6, 2, 1, 1, 6, 7, 4, 7, 2, 8, 8, 1, 8, 7, 1, 5, 3, 8, 6, 3, 9, 6, 7, 7, 5, 2, 6, 5, 6, 6, 7, 4, 9, 2, 5, 3, 8, 5, 2, 9, 3, 4, 4, 5, 2, 7, 3, 6, 5, 2, 6, 5, 6, 8, 5, 5, 3, 3, 2, 4, 1, 2, 2, 7, 6, 7, 4, 1, 8, 4, 3, 6, 8, 4, 5, 8, 7, 8, 4, 7, 4, 8, 2, 4, 8, 9, 9, 5, 6, 4, 7, 7, 3, 4, 6, 8, 1, 8, 2, 3, 6, 1, 6, 9, 9, 3, 6, 4, 2, 8, 4, 9, 6, 6, 2, 8, 3, 2, 1, 2, 1, 2, 2, 7, 9, 9, 4, 3, 1, 8, 5, 7, 1, 6, 4, 4, 5, 1, 7, 6, 4, 1, 1, 2, 3, 1, 3, 3, 7, 8, 9, 9, 3, 5, 8, 8, 6, 8, 3, 5, 2, 2, 2, 8, 1, 1, 3, 2, 5, 7, 1, 7, 7, 3, 9, 7, 7, 6, 3, 6, 1, 4, 8, 1, 5, 2, 4, 6, 8, 7
\end{enumerate}
}

\vspace{0.5cm}

\subsubsection*{\large AIQ Numeric Manipulation}

\noindent\textbf{Example A. (easy item):} \textit{“Which one of the answer options below (A, B, C, or D) depicts the following string of numbers in reverse order?"  5, 1, 1, 8, 1, 1}
\begin{enumerate}[label=\Alph*., leftmargin=2em]
    \item 1, 1, 1, 8, 1, 5
    \item 1, 1, 1, 1, 8, 5
    \item \textbf{1, 1, 8, 1, 1, 5 \textcolor{darkgreen}{(GT)}}
    \item 1, 1, 8, 5, 1, 1
\end{enumerate}

\vspace{0.3cm}

\noindent\textbf{Example B. (difficult item):} \textit{“Which one of the answer options below (A, B, C, or D) depicts the following string of numbers in reverse order?”  7, 6, 3, 8, 6, 7, 4, 6, 2, 9, 5, 9, 1, 3, 3, 8, 6, 1, 8, 3, 4, 7, 9, 3, 1, 5, 2, 9, 9, 2, 6, 6, 1, 5, 3, 6, 7, 6, 6, 1, 6, 5, 4, 8, 3, 9, 6, 7, 7, 9, 9, 4, 4, 1, 3, 5, 3, 2, 8, 6, 2, 4, 8, 5, 2, 7, 6, 4, 4, 7, 3, 6, 8, 5, 1, 9, 3, 7, 7, 5, 9, 9, 5, 7, 5, 6, 6, 8, 6, 4, 5, 6, 7, 4, 3, 6, 7, 4, 5, 1, 2, 7, 6, 6, 6, 9, 5, 7, 2, 4, 1, 5, 7, 7, 2, 5, 1, 1, 1, 1, 6, 6, 5, 4, 1, 7, 1, 1, 2, 4, 6, 8, 5, 9, 8, 2, 8, 2, 7, 3, 8, 4, 8, 6, 9, 6, 2, 6, 1, 6, 9, 7, 5, 2, 3, 5, 1, 8, 9, 5, 7, 6, 9, 2, 8, 5, 8, 4, 1, 2, 1, 9, 5, 1, 8, 8, 7, 3, 6, 1, 4, 8, 3, 4, 9, 7, 7, 8, 1, 2, 2, 9, 2, 2, 9, 6, 7, 3, 3, 9, 7, 6, 1, 8, 6, 7, 9, 3, 1, 9, 2, 3, 7, 4, 6, 1, 4, 9, 3, 9, 4, 5, 6, 5, 5, 9, 3, 8, 3, 9, 3, 6, 2, 9, 2, 1, 1, 1, 8, 7, 4, 1, 1, 6, 7, 6, 8, 4, 2, 4, 1, 9, 4, 7, 6, 9, 8, 1, 1, 8, 2, 8, 7, 6, 5, 5, 6, 4, 7, 1, 1, 5, 7, 7, 1, 5, 6, 8, 6, 3, 1, 3, 5, 7, 1, 8, 9, 7, 7, 4, 8, 6, 8, 2, 2, 4, 2, 5, 7, 6, 3, 2, 2, 7, 8, 8, 7, 6, 4, 7, 9, 9, 7, 8, 8, 6, 5, 8, 9, 6, 4, 6, 1, 5, 8, 2, 4, 8, 5, 6, 1, 6, 7, 8, 7, 1, 9, 3, 9, 8, 2, 4, 9, 6, 2, 1, 5, 4, 2, 2, 6, 7, 5, 6, 8, 7, 1, 1, 6, 3, 9, 8, 2, 9, 2, 5, 6, 2, 6, 5, 6, 2, 6, 1, 5, 3, 1, 4, 3, 6, 1, 7, 8, 4, 6, 9, 1, 7, 7, 6, 9, 2, 3, 7, 7, 2, 6, 6, 5, 7}

{\raggedright \small
\begin{enumerate}[label=\Alph*., leftmargin=*, itemsep=1ex]
    \item 7, 5, 6, 6, 2, 7, 7, 3, 2, 9, 6, 7, 7, 1, 9, 6, 4, 8, 7, 1, 6, 3, 4, 1, 3, 5, 1, 6, 2, 6, 5, 6, 2, 6, 5, 2, 9, 2, 8, 9, 3, 6, 1, 1, 7, 8, 6, 5, 7, 6, 2, 2, 4, 5, 1, 2, 6, 9, 4, 2, 8, 9, 3, 9, 1, 7, 8, 7, 6, 1, 6, 5, 8, 4, 2, 8, 5, 1, 6, 4, 6, 9, 8, 5, 6, 8, 8, 7, 9, 9, 7, 4, 6, 7, 8, 8, 7, 2, 2, 3, 6, 7, 5, 2, 4, 2, 2, 8, 6, 8, 4, 7, 7, 9, 8, 1, 7, 5, 3, 1, 3, 6, 8, 6, 5, 1, 7, 7, 5, 1, 1, 7, 6, 6, 5, 5, 6, 7, 8, 2, 8, 1, 1, 8, 9, 6, 7, 4, 9, 1, 4, 2, 4, 8, 6, 7, 6, 1, 1, 4, 7, 8, 1, 1, 1, 2, 9, 2, 6, 3, 9, 3, 8, 3, 9, 5, 5, 6, 5, 4, 9, 3, 9, 4, 1, 6, 4, 7, 3, 2, 9, 1, 3, 9, 7, 6, 8, 1, 6, 7, 9, 3, 3, 7, 6, 9, 2, 2, 9, 2, 2, 1, 8, 7, 7, 9, 4, 3, 8, 4, 1, 6, 3, 7, 8, 8, 1, 5, 9, 1, 2, 1, 4, 8, 5, 8, 2, 9, 6, 7, 5, 9, 8, 1, 5, 3, 2, 5, 7, 9, 6, 1, 6, 2, 6, 9, 6, 8, 4, 8, 3, 7, 2, 8, 2, 8, 9, 5, 8, 6, 4, 2, 1, 1, 7, 1, 4, 5, 6, 6, 1, 1, 1, 1, 5, 2, 7, 7, 5, 1, 4, 2, 7, 5, 9, 6, 6, 6, 7, 2, 1, 5, 4, 7, 6, 3, 4, 7, 6, 5, 4, 6, 8, 4, 6, 5, 7, 5, 9, 9, 5, 7, 7, 3, 9, 1, 5, 8, 6, 3, 7, 4, 4, 6, 7, 2, 5, 8, 4, 2, 6, 8, 2, 3, 5, 3, 1, 4, 4, 9, 9, 7, 7, 6, 9, 3, 8, 4, 5, 6, 1, 6, 6, 7, 6, 3, 5, 1, 6, 6, 2, 9, 9, 2, 5, 1, 3, 9, 7, 4, 3, 8, 1, 6, 8, 3, 3, 1, 9, 5, 9, 2, 6, 4, 7, 6, 8, 3, 6, 7
    \item 7, 5, 6, 6, 2, 7, 7, 3, 2, 9, 6, 7, 7, 1, 9, 6, 4, 8, 7, 1, 6, 3, 4, 1, 3, 5, 1, 6, 2, 6, 5, 6, 2, 6, 5, 2, 9, 2, 8, 9, 3, 6, 1, 1, 7, 8, 6, 5, 7, 6, 2, 2, 4, 5, 1, 2, 6, 9, 4, 2, 8, 9, 3, 9, 1, 7, 8, 7, 6, 1, 6, 5, 8, 4, 2, 8, 5, 1, 6, 4, 6, 9, 8, 5, 6, 8, 8, 7, 9, 9, 7, 4, 6, 7, 8, 8, 7, 2, 2, 3, 6, 7, 5, 2, 4, 2, 2, 8, 6, 8, 4, 7, 7, 9, 8, 1, 7, 5, 3, 1, 3, 6, 8, 6, 5, 1, 7, 7, 5, 1, 1, 7, 4, 6, 5, 5, 6, 7, 8, 2, 8, 1, 1, 8, 9, 6, 7, 4, 9, 1, 4, 2, 4, 8, 6, 7, 6, 1, 1, 4, 7, 8, 1, 1, 8, 2, 9, 2, 6, 3, 9, 3, 8, 3, 9, 5, 5, 6, 5, 4, 9, 3, 9, 4, 1, 6, 4, 7, 3, 2, 9, 1, 3, 9, 7, 6, 8, 1, 6, 7, 9, 3, 3, 7, 6, 9, 2, 2, 9, 2, 2, 1, 8, 7, 7, 9, 4, 3, 8, 4, 1, 6, 3, 7, 8, 8, 1, 5, 9, 1, 2, 1, 4, 8, 5, 8, 2, 9, 6, 7, 5, 9, 8, 1, 5, 3, 2, 5, 7, 9, 6, 1, 6, 2, 6, 9, 6, 8, 4, 8, 3, 7, 2, 8, 2, 8, 9, 5, 8, 6, 4, 2, 1, 1, 7, 1, 4, 5, 6, 6, 1, 1, 1, 1, 5, 2, 7, 7, 5, 1, 4, 2, 7, 5, 9, 6, 6, 6, 7, 2, 1, 5, 4, 7, 6, 3, 4, 7, 6, 5, 4, 6, 8, 6, 6, 5, 7, 5, 9, 9, 5, 7, 7, 3, 9, 1, 5, 8, 6, 3, 7, 4, 4, 6, 7, 2, 5, 8, 4, 2, 6, 8, 2, 3, 5, 3, 1, 4, 4, 9, 9, 7, 7, 6, 9, 3, 8, 4, 5, 6, 1, 6, 6, 7, 6, 3, 5, 1, 6, 6, 2, 9, 9, 2, 5, 1, 3, 9, 7, 4, 3, 8, 1, 6, 1, 3, 3, 1, 9, 5, 9, 2, 6, 4, 7, 6, 8, 3, 6, 7
    \item 7, 5, 6, 6, 2, 7, 7, 3, 2, 9, 6, 7, 7, 1, 9, 6, 4, 8, 7, 1, 6, 3, 4, 1, 3, 5, 1, 6, 2, 6, 5, 6, 2, 6, 5, 2, 9, 2, 8, 9, 3, 6, 1, 1, 7, 8, 6, 5, 7, 6, 2, 2, 4, 5, 1, 2, 6, 9, 4, 2, 8, 9, 3, 9, 1, 7, 8, 7, 6, 1, 6, 5, 8, 4, 2, 8, 5, 1, 6, 4, 6, 9, 8, 5, 6, 8, 8, 7, 9, 9, 7, 4, 6, 7, 8, 8, 7, 2, 2, 3, 6, 7, 5, 2, 4, 2, 2, 8, 6, 8, 4, 7, 7, 9, 8, 1, 7, 5, 3, 1, 3, 6, 8, 6, 5, 1, 7, 7, 5, 1, 1, 7, 4, 6, 5, 5, 6, 7, 8, 2, 8, 1, 1, 8, 9, 6, 7, 4, 9, 1, 4, 2, 4, 8, 6, 7, 6, 1, 1, 4, 7, 8, 1, 1, 1, 2, 9, 2, 6, 3, 9, 3, 8, 3, 9, 5, 5, 6, 5, 4, 9, 3, 9, 4, 1, 6, 4, 7, 3, 2, 9, 1, 3, 9, 7, 6, 8, 1, 6, 7, 9, 3, 3, 7, 6, 9, 2, 2, 9, 2, 2, 1, 8, 7, 7, 9, 4, 3, 8, 4, 1, 6, 3, 7, 8, 8, 1, 5, 9, 1, 2, 1, 4, 8, 5, 8, 2, 9, 3, 7, 5, 9, 8, 1, 5, 3, 2, 5, 7, 9, 6, 1, 6, 2, 6, 9, 6, 8, 4, 8, 3, 7, 2, 8, 2, 8, 9, 5, 8, 6, 4, 2, 1, 1, 7, 1, 4, 5, 6, 6, 1, 1, 1, 1, 5, 2, 7, 7, 5, 1, 4, 2, 7, 5, 9, 6, 6, 6, 7, 2, 1, 5, 4, 7, 6, 3, 4, 7, 6, 5, 4, 6, 8, 6, 6, 5, 7, 5, 9, 9, 5, 7, 7, 6, 9, 1, 5, 8, 6, 3, 7, 4, 4, 6, 7, 2, 5, 8, 4, 2, 6, 8, 2, 3, 5, 3, 1, 4, 4, 9, 9, 7, 7, 6, 9, 3, 8, 4, 5, 6, 1, 6, 6, 7, 6, 3, 5, 1, 6, 6, 2, 9, 9, 2, 5, 1, 3, 9, 7, 4, 3, 8, 1, 6, 8, 3, 3, 1, 9, 5, 9, 2, 6, 4, 7, 6, 8, 3, 6, 7
    \item \textbf{7, 5, 6, 6, 2, 7, 7, 3, 2, 9, 6, 7, 7, 1, 9, 6, 4, 8, 7, 1, 6, 3, 4, 1, 3, 5, 1, 6, 2, 6, 5, 6, 2, 6, 5, 2, 9, 2, 8, 9, 3, 6, 1, 1, 7, 8, 6, 5, 7, 6, 2, 2, 4, 5, 1, 2, 6, 9, 4, 2, 8, 9, 3, 9, 1, 7, 8, 7, 6, 1, 6, 5, 8, 4, 2, 8, 5, 1, 6, 4, 6, 9, 8, 5, 6, 8, 8, 7, 9, 9, 7, 4, 6, 7, 8, 8, 7, 2, 2, 3, 6, 7, 5, 2, 4, 2, 2, 8, 6, 8, 4, 7, 7, 9, 8, 1, 7, 5, 3, 1, 3, 6, 8, 6, 5, 1, 7, 7, 5, 1, 1, 7, 4, 6, 5, 5, 6, 7, 8, 2, 8, 1, 1, 8, 9, 6, 7, 4, 9, 1, 4, 2, 4, 8, 6, 7, 6, 1, 1, 4, 7, 8, 1, 1, 1, 2, 9, 2, 6, 3, 9, 3, 8, 3, 9, 5, 5, 6, 5, 4, 9, 3, 9, 4, 1, 6, 4, 7, 3, 2, 9, 1, 3, 9, 7, 6, 8, 1, 6, 7, 9, 3, 3, 7, 6, 9, 2, 2, 9, 2, 2, 1, 8, 7, 7, 9, 4, 3, 8, 4, 1, 6, 3, 7, 8, 8, 1, 5, 9, 1, 2, 1, 4, 8, 5, 8, 2, 9, 6, 7, 5, 9, 8, 1, 5, 3, 2, 5, 7, 9, 6, 1, 6, 2, 6, 9, 6, 8, 4, 8, 3, 7, 2, 8, 2, 8, 9, 5, 8, 6, 4, 2, 1, 1, 7, 1, 4, 5, 6, 6, 1, 1, 1, 1, 5, 2, 7, 7, 5, 1, 4, 2, 7, 5, 9, 6, 6, 6, 7, 2, 1, 5, 4, 7, 6, 3, 4, 7, 6, 5, 4, 6, 8, 6, 6, 5, 7, 5, 9, 9, 5, 7, 7, 3, 9, 1, 5, 8, 6, 3, 7, 4, 4, 6, 7, 2, 5, 8, 4, 2, 6, 8, 2, 3, 5, 3, 1, 4, 4, 9, 9, 7, 7, 6, 9, 3, 8, 4, 5, 6, 1, 6, 6, 7, 6, 3, 5, 1, 6, 6, 2, 9, 9, 2, 5, 1, 3, 9, 7, 4, 3, 8, 1, 6, 8, 3, 3, 1, 9, 5, 9, 2, 6, 4, 7, 6, 8, 3, 6, 7 \textcolor{darkgreen}{(GT)}}
\end{enumerate}
}

\vspace{0.5cm}

\subsubsection*{\large AIQ Numeric Sequencing}

\noindent\textbf{Example A. (easy item):} \textit{“Which one of the answer options below (A, B, C, or D) depicts the following string of numbers in order from lowest to highest?” 1, 6, 8, 6, 4, 2}
\begin{enumerate}[label=\Alph*., leftmargin=2em]
    \item 1, 6, 4, 2, 6, 8
    \item \textbf{1, 2, 4, 6, 6, 8 \textcolor{darkgreen}{(GT)}}
    \item 1, 2, 4, 8, 6, 6
    \item 4, 2, 1, 6, 6, 8
\end{enumerate}

\vspace{0.3cm}

\noindent\textbf{Example B. (difficult item):} \textit{“Which one of the answer options below (A, B, C, or D) depicts the following string of numbers in order from lowest to highest?” 2, 9, 1, 6, 7, 2, 3, 2, 5, 9, 6, 7, 7, 5, 9, 2, 9, 5, 6, 3, 5, 8, 5, 9, 9, 8, 1, 4, 5, 6, 8, 5, 1, 1, 2, 3, 4, 3, 7, 6, 7, 2, 8, 5, 4, 8, 3, 4, 2, 1, 1, 3, 6, 9, 5, 2, 7, 5, 4, 4, 5, 6, 6, 6, 1, 4, 2, 2, 9, 2, 3, 7, 8, 7, 8, 9, 8, 2, 7, 5, 5, 6, 8, 7, 5, 1, 6, 8, 7, 9, 9, 9, 2, 9, 4, 6, 9, 8, 8, 9, 7, 5, 5, 4, 7, 5, 8, 9, 4, 1, 8, 6, 2, 7, 5, 9, 4, 9, 2, 2, 5, 2, 3, 3, 6, 7, 3, 4, 5, 4, 1, 6, 5, 4, 3, 4, 3, 1, 5, 2, 6, 4, 3, 2, 7, 1, 5, 1, 7, 8, 7, 1, 4, 5, 2, 1, 7, 6, 6, 4, 4, 2, 9, 7, 6, 3, 5, 3, 8, 6, 8, 7, 7, 2, 8, 2, 5, 4, 7, 4, 7, 2, 6, 6, 9, 9, 2, 6, 1, 3, 9, 5, 8, 1, 2, 2, 9, 6, 7, 2, 5, 9, 2, 5, 2, 5, 1, 9, 2, 3, 3, 5, 8, 2, 4, 6, 5, 9, 4, 9, 8, 4, 3, 8, 9, 4, 1, 2, 4, 3, 5, 8, 7, 2, 8, 3, 5, 1, 1, 7, 8, 6, 5, 8, 8, 9, 2, 5, 6, 7, 1, 3, 6, 7, 2, 4, 5, 7, 3, 8, 4, 7, 6, 1, 8, 7, 6, 1, 5, 5, 9, 6, 9, 5, 2, 6, 3, 2, 9, 5, 5, 3, 6, 3, 8, 6, 8, 1, 3, 4, 9, 7, 5, 2, 5, 6, 5, 5, 3, 1, 3, 8, 5, 6, 9, 7, 7, 5, 2, 7, 5, 5, 9, 5, 9, 3, 4, 3, 3, 7, 5, 9, 1, 7, 6, 9, 1, 8, 9, 5, 2, 8, 6, 1, 2, 6, 1, 9, 1, 8, 9, 2, 7, 2, 8, 6, 7, 4, 1, 7, 8, 8, 3, 6, 9, 3, 9, 3, 5, 8, 6, 3, 7, 1, 6, 2, 5, 1, 8, 5, 8, 2, 9, 5, 3, 8, 4, 3, 8, 8, 2, 2, 5, 6, 3, 7, 3, 5, 2, 8, 5, 6, 5, 9, 4, 7, 8, 3, 4, 7}

{\raggedright \small
\begin{enumerate}[label=\Alph*., leftmargin=*, itemsep=1ex]
    \item 1, 1, 1, 1, 1, 1, 1, 1, 1, 1, 1, 1, 1, 1, 1, 1, 1, 1, 1, 4, 1, 1, 1, 1, 1, 1, 1, 1, 1, 1, 1, 1, 1, 1, 2, 2, 2, 2, 2, 2, 2, 2, 2, 2, 2, 2, 2, 2, 2, 2, 2, 2, 2, 2, 2, 2, 2, 2, 2, 2, 2, 2, 2, 2, 2, 2, 2, 2, 2, 2, 2, 2, 2, 2, 2, 2, 2, 2, 2, 2, 2, 2, 2, 3, 3, 3, 3, 3, 3, 3, 3, 3, 3, 3, 3, 3, 3, 3, 3, 3, 3, 3, 3, 3, 3, 3, 3, 3, 3, 3, 3, 3, 3, 3, 3, 3, 3, 3, 3, 3, 3, 3, 3, 3, 4, 4, 4, 4, 4, 4, 4, 4, 4, 4, 4, 1, 4, 4, 4, 4, 4, 4, 4, 4, 4, 4, 4, 4, 4, 4, 4, 4, 4, 4, 4, 4, 4, 4, 5, 5, 5, 5, 5, 5, 5, 5, 5, 5, 5, 5, 5, 5, 5, 5, 5, 5, 5, 5, 5, 5, 5, 5, 5, 5, 5, 5, 5, 5, 5, 5, 5, 5, 5, 5, 5, 5, 5, 5, 5, 5, 5, 5, 5, 5, 5, 5, 5, 5, 5, 5, 5, 5, 5, 5, 5, 5, 5, 5, 5, 6, 6, 6, 6, 6, 6, 6, 6, 6, 6, 6, 6, 6, 6, 6, 6, 6, 6, 6, 6, 6, 6, 6, 6, 6, 6, 6, 6, 6, 6, 6, 6, 6, 6, 6, 6, 6, 6, 6, 6, 6, 6, 6, 6, 6, 7, 7, 7, 7, 7, 7, 7, 7, 7, 7, 7, 7, 7, 7, 7, 7, 7, 7, 7, 7, 7, 7, 7, 7, 7, 7, 7, 7, 7, 7, 7, 7, 7, 7, 7, 7, 7, 7, 7, 7, 7, 7, 7, 7, 7, 8, 8, 8, 8, 8, 8, 8, 8, 8, 8, 8, 8, 8, 8, 8, 8, 8, 8, 8, 8, 8, 8, 8, 8, 8, 8, 8, 8, 8, 8, 8, 8, 8, 8, 8, 8, 8, 8, 8, 8, 8, 8, 8, 8, 8, 8, 9, 9, 9, 9, 9, 9, 9, 9, 9, 9, 9, 9, 9, 9, 9, 9, 9, 9, 9, 9, 9, 9, 9, 9, 9, 9, 9, 9, 9, 9, 9, 9, 9, 9, 9, 9, 9, 9, 9, 9, 9, 9, 9, 9, 9
    \item 1, 1, 1, 1, 1, 1, 1, 1, 1, 1, 1, 1, 1, 1, 1, 1, 1, 1, 1, 1, 1, 1, 1, 1, 1, 1, 1, 1, 1, 1, 1, 1, 1, 1, 2, 2, 2, 2, 2, 2, 2, 2, 2, 2, 2, 2, 2, 2, 2, 2, 2, 2, 2, 2, 2, 2, 2, 2, 2, 2, 2, 2, 2, 2, 2, 2, 2, 2, 2, 2, 2, 2, 2, 2, 2, 2, 2, 2, 2, 2, 2, 2, 2, 3, 3, 3, 3, 3, 3, 3, 3, 3, 3, 3, 3, 3, 3, 3, 3, 3, 3, 3, 3, 3, 3, 3, 3, 3, 3, 3, 3, 3, 3, 3, 3, 3, 3, 3, 3, 3, 3, 3, 3, 3, 4, 4, 4, 4, 4, 4, 4, 4, 4, 4, 4, 4, 4, 4, 4, 4, 4, 4, 4, 4, 4, 4, 4, 4, 4, 4, 4, 4, 4, 4, 4, 4, 4, 4, 5, 5, 5, 5, 5, 5, 5, 5, 5, 5, 5, 5, 5, 5, 5, 5, 5, 5, 5, 5, 5, 5, 5, 5, 5, 5, 5, 5, 5, 5, 5, 5, 5, 5, 5, 5, 5, 5, 5, 5, 5, 5, 5, 5, 5, 5, 5, 5, 5, 5, 5, 5, 5, 5, 5, 5, 5, 5, 5, 5, 5, 6, 6, 6, 6, 6, 6, 6, 6, 6, 6, 6, 6, 6, 6, 6, 6, 6, 6, 6, 6, 6, 6, 6, 6, 6, 6, 6, 6, 6, 6, 6, 6, 6, 7, 6, 6, 6, 6, 6, 6, 6, 6, 6, 6, 6, 7, 7, 7, 7, 7, 7, 7, 7, 7, 7, 7, 7, 7, 7, 7, 7, 7, 7, 7, 7, 7, 7, 7, 7, 7, 7, 7, 7, 7, 7, 7, 7, 7, 7, 7, 7, 7, 6, 7, 7, 7, 7, 7, 7, 7, 8, 8, 8, 8, 8, 8, 8, 8, 8, 8, 8, 8, 8, 8, 8, 8, 8, 8, 8, 8, 8, 8, 8, 8, 8, 8, 8, 8, 8, 8, 8, 8, 8, 8, 8, 8, 8, 8, 8, 8, 8, 8, 8, 8, 8, 8, 9, 9, 9, 9, 9, 9, 9, 9, 9, 9, 9, 9, 9, 9, 9, 9, 9, 9, 9, 9, 9, 9, 9, 9, 9, 9, 9, 9, 9, 9, 9, 9, 9, 9, 9, 9, 9, 9, 9, 9, 9, 9, 9, 9, 9
    \item 1, 1, 1, 1, 1, 1, 1, 1, 1, 1, 1, 1, 1, 1, 1, 1, 1, 1, 1, 1, 1, 1, 1, 1, 1, 1, 1, 1, 1, 1, 1, 1, 1, 1, 2, 2, 2, 2, 2, 2, 2, 2, 2, 2, 2, 2, 2, 2, 2, 2, 2, 2, 2, 2, 2, 2, 2, 2, 2, 2, 2, 2, 2, 2, 2, 2, 2, 2, 2, 2, 2, 2, 2, 2, 2, 2, 2, 2, 2, 2, 2, 2, 2, 3, 3, 3, 3, 3, 3, 3, 3, 3, 3, 3, 3, 3, 3, 3, 3, 3, 3, 3, 3, 3, 3, 3, 3, 3, 3, 3, 3, 3, 3, 3, 3, 3, 3, 3, 3, 3, 3, 3, 3, 3, 4, 4, 4, 4, 4, 4, 4, 4, 4, 4, 4, 4, 4, 4, 4, 4, 4, 4, 4, 4, 4, 4, 4, 4, 4, 4, 4, 4, 4, 4, 4, 4, 4, 4, 5, 5, 5, 5, 5, 5, 5, 5, 5, 5, 5, 5, 5, 5, 5, 5, 5, 5, 5, 5, 5, 5, 5, 5, 5, 5, 5, 5, 5, 5, 5, 5, 5, 5, 5, 6, 5, 5, 5, 5, 5, 5, 5, 5, 5, 5, 5, 5, 5, 5, 5, 5, 5, 5, 5, 5, 5, 5, 5, 5, 5, 6, 6, 6, 6, 6, 6, 6, 6, 6, 6, 6, 6, 6, 6, 6, 6, 6, 6, 6, 6, 6, 6, 6, 6, 6, 6, 6, 6, 6, 6, 6, 6, 6, 6, 6, 6, 5, 6, 6, 6, 6, 6, 6, 6, 6, 7, 7, 7, 7, 7, 7, 7, 7, 7, 7, 7, 7, 7, 7, 7, 7, 7, 7, 7, 7, 7, 7, 7, 7, 7, 7, 7, 7, 7, 7, 7, 7, 7, 7, 7, 7, 7, 7, 7, 7, 7, 7, 7, 7, 7, 8, 8, 8, 8, 8, 8, 8, 8, 8, 8, 8, 8, 8, 8, 8, 8, 8, 8, 8, 8, 8, 8, 8, 8, 8, 8, 8, 8, 8, 8, 8, 8, 8, 8, 8, 8, 8, 8, 8, 8, 8, 8, 8, 8, 8, 8, 9, 9, 9, 9, 9, 9, 9, 9, 9, 9, 9, 9, 9, 9, 9, 9, 9, 9, 9, 9, 9, 9, 9, 9, 9, 9, 9, 9, 9, 9, 9, 9, 9, 9, 9, 9, 9, 9, 9, 9, 9, 9, 9, 9, 9
    \item \textbf{1, 1, 1, 1, 1, 1, 1, 1, 1, 1, 1, 1, 1, 1, 1, 1, 1, 1, 1, 1, 1, 1, 1, 1, 1, 1, 1, 1, 1, 1, 1, 1, 1, 1, 2, 2, 2, 2, 2, 2, 2, 2, 2, 2, 2, 2, 2, 2, 2, 2, 2, 2, 2, 2, 2, 2, 2, 2, 2, 2, 2, 2, 2, 2, 2, 2, 2, 2, 2, 2, 2, 2, 2, 2, 2, 2, 2, 2, 2, 2, 2, 2, 2, 3, 3, 3, 3, 3, 3, 3, 3, 3, 3, 3, 3, 3, 3, 3, 3, 3, 3, 3, 3, 3, 3, 3, 3, 3, 3, 3, 3, 3, 3, 3, 3, 3, 3, 3, 3, 3, 3, 3, 3, 3, 4, 4, 4, 4, 4, 4, 4, 4, 4, 4, 4, 4, 4, 4, 4, 4, 4, 4, 4, 4, 4, 4, 4, 4, 4, 4, 4, 4, 4, 4, 4, 4, 4, 4, 5, 5, 5, 5, 5, 5, 5, 5, 5, 5, 5, 5, 5, 5, 5, 5, 5, 5, 5, 5, 5, 5, 5, 5, 5, 5, 5, 5, 5, 5, 5, 5, 5, 5, 5, 5, 5, 5, 5, 5, 5, 5, 5, 5, 5, 5, 5, 5, 5, 5, 5, 5, 5, 5, 5, 5, 5, 5, 5, 5, 5, 6, 6, 6, 6, 6, 6, 6, 6, 6, 6, 6, 6, 6, 6, 6, 6, 6, 6, 6, 6, 6, 6, 6, 6, 6, 6, 6, 6, 6, 6, 6, 6, 6, 6, 6, 6, 6, 6, 6, 6, 6, 6, 6, 6, 6, 7, 7, 7, 7, 7, 7, 7, 7, 7, 7, 7, 7, 7, 7, 7, 7, 7, 7, 7, 7, 7, 7, 7, 7, 7, 7, 7, 7, 7, 7, 7, 7, 7, 7, 7, 7, 7, 7, 7, 7, 7, 7, 7, 7, 7, 8, 8, 8, 8, 8, 8, 8, 8, 8, 8, 8, 8, 8, 8, 8, 8, 8, 8, 8, 8, 8, 8, 8, 8, 8, 8, 8, 8, 8, 8, 8, 8, 8, 8, 8, 8, 8, 8, 8, 8, 8, 8, 8, 8, 8, 8, 9, 9, 9, 9, 9, 9, 9, 9, 9, 9, 9, 9, 9, 9, 9, 9, 9, 9, 9, 9, 9, 9, 9, 9, 9, 9, 9, 9, 9, 9, 9, 9, 9, 9, 9, 9, 9, 9, 9, 9, 9, 9, 9, 9, 9 \textcolor{darkgreen}{(GT)}}
\end{enumerate}
}

\vspace{0.5cm}

\subsubsection*{\large AIQ Alphanumeric Sequencing}

\noindent\textbf{Example A. (easy item):} \textit{"Which one of the answer options below (A, B, C, or D) depicts the following string properly sequenced, with numbers first (ordered from lowest to highest), followed by letters in alphabetical order?”  Q, F, H, F, 8, 4, 1, Q, R, 5}
\begin{enumerate}[label=\Alph*., leftmargin=2em]
    \item 5, 4, 1, 8, F, F, H, Q, Q, R
    \item 4, 1, 5, 8, F, F, H, Q, Q, R
    \item \textbf{1, 4, 5, 8, F, F, H, Q, Q, R \textcolor{darkgreen}{(GT)}}
    \item 1, 4, F, 8, 5, F, H, Q, Q, R
\end{enumerate}

\vspace{0.3cm}

\noindent\textbf{Example B. (difficult item):} \textit{"Which one of the answer options below (A, B, C, or D) depicts the following string properly sequenced, with numbers first (ordered from lowest to highest), followed by letters in alphabetical order?”
1, Q, W, A, M, 3, Z, T, O, C, L, B, C, 5, L, F, 4, H, A, 6, R, C, 1, A, T, B, J, O, 9, C, G, E, 5, C, X, C, A, B, U, N, F, T, O, 7, H, X, V, 6, V, A, F, 9, 8, H, C, F, T, X, T, P, S, I, G, N, L, 4, 3, S, Z, 9, J, 2, P, 7, 2, I, G, 3, 9, V, C, 8, Z, F, 2, 9, 7, F, L, G, G, I, C, B, J, 4, W, 2, 3, D, 6, L, Z, R, Q, 9, H, W, P, B, S, N, U, H, M, K, O, 4, S, 1, J, W, H, N, 6, E, W, 5, J, M, X, 6, G, B, Q, S, L, 2, 9, D, C, C, V, T, D, Q, H, X, I, S, J, 7, 2, N, S, X, B, V, W, 3, L, Y, P, C, X, Z, 8, 1, X, M, O, R, T, Q, F, F, P, O, H, F, 7, 8, L, T, Z, S, Y, Q, T, V, 3, Q, G, U, 6, O, 6, 6, 8, Q, A, Y, R, M, 8, W, N, E, O, R, O, L, W, Y, C, 9, O, 4, M, Y, 5, Z, 5, M, T, W, H, R, B, Q, W, O, A, X, A, A, 7, 9, T, 8, U, F, C, L, 1, 3, N, 8, B, S, H, X, 3, 5, Z, Y, 4, O, L, K, K, D, Q, Y, 8, K, C, O, C, 3, E, B, L, D, E, B, E, A, 2, 3, Z, R, Z, 7, P, F, L, 2, 9, D, E, V, S, W, X, V, M, J, 8, 5, C, R, V, 8, J, Q, 8, 6, G, V, 3, 7, S, E, Y, B, E, D, X, P, T, K, G, D, D, Z, K, U, C, P, 7, A, Y, W, J, 5, Z, B, 8, I, Y, F, S, X, R, F, A, M, H, 6, G, 3, L, T, 7, O, 1, H, F, E, H, U, F, E, Q, F, Y, M, 9, Y, G, Z, U, I, Y, O, 3, C, F, 4, O, L, F, I, V, U, T, P, I, 4, R, O, 3, R, D, Z, M, P, 4, 8}

{\raggedright \small
\begin{enumerate}[label=\Alph*., leftmargin=*, itemsep=1ex]
    \item 1, 1, 1, 1, 1, 1, 2, 2, 2, 2, 5, 2, 2, 2, 3, 3, 3, 3, 3, 3, 3, 3, 3, 3, 3, 3, 3, 3, 4, 4, 4, 4, 4, 4, 4, 4, 4, 5, 5, 5, 5, 5, 5, 2, 5, 6, 6, 6, 6, 6, 6, 6, 6, 6, 6, 7, 7, 7, 7, 7, 7, 7, 7, 7, 7, 8, 8, 8, 8, 8, 8, 8, 8, 8, 8, 8, 8, 8, 8, 9, 9, 9, 9, 9, 9, 9, 9, 9, 9, 9, A, A, A, A, A, A, A, A, A, A, A, A, B, B, B, B, B, B, B, B, B, B, B, B, B, C, C, C, C, C, C, C, C, C, C, C, C, C, C, C, C, C, C, C, D, D, D, D, D, D, D, D, D, D, E, E, E, E, E, E, E, E, E, E, E, F, F, F, F, F, F, F, F, F, F, F, F, F, F, F, F, F, F, G, G, G, G, G, G, G, G, G, G, G, H, H, H, H, H, H, H, H, H, H, H, H, H, I, I, I, I, I, I, I, I, J, J, J, J, J, J, J, J, J, K, K, K, K, K, K, L, L, L, L, L, L, L, L, L, L, L, L, L, L, L, M, M, M, M, M, M, M, M, M, M, M, N, N, N, N, N, N, N, O, O, O, O, O, O, O, O, O, O, O, O, O, O, O, O, O, P, P, P, P, P, P, P, P, P, P, Q, Q, Q, Q, Q, Q, Q, Q, Q, Q, Q, Q, R, R, R, R, R, R, R, R, R, R, R, S, S, S, S, S, S, S, S, S, S, S, S, T, T, T, T, T, T, T, T, T, T, T, T, T, T, U, U, U, U, U, U, U, U, V, V, V, V, V, V, V, V, V, V, V, W, W, W, W, W, W, W, W, W, W, W, W, X, X, X, X, X, X, X, X, X, X, X, X, X, Y, Y, Y, Y, Y, Y, Y, Y, Y, Y, Y, Y, Y, Z, Z, Z, Z, Z, Z, Z, Z, Z, Z, Z, Z, Z, Z
    \item 1, 1, 1, 1, 1, 1, 2, 2, 2, 2, 2, 2, 2, 2, 3, 3, 3, 3, 3, 3, 3, 3, 3, 3, 3, 3, 3, 3, 4, 4, 4, 4, 4, 4, 4, 4, 4, 5, 5, 5, 5, 5, 5, 5, 5, 6, 6, 6, 6, 6, 6, 6, 6, 6, 6, 7, 7, 7, 7, 7, 7, 7, 7, 7, 7, 8, 8, 8, 8, 8, 8, 8, 8, 8, 8, 8, 8, 8, 8, 9, 9, 9, 9, 9, 9, 9, 9, 9, 9, 9, A, A, A, A, A, A, A, A, A, A, A, A, B, B, B, B, B, B, B, B, B, B, B, B, B, C, C, C, C, C, C, C, C, E, C, C, C, C, C, C, C, C, C, C, D, D, D, D, D, D, D, D, D, D, C, E, E, E, E, E, E, E, E, E, E, F, F, F, F, F, F, F, F, F, F, F, F, F, F, F, F, F, F, G, G, G, G, G, G, G, G, G, G, G, H, H, H, H, H, H, H, H, H, H, H, H, H, I, I, I, I, I, I, I, I, J, J, J, J, J, J, J, J, J, K, K, K, K, K, K, L, L, L, L, L, L, L, L, L, L, L, L, L, L, L, M, M, M, M, M, M, M, M, M, M, M, N, N, N, N, N, N, N, O, O, O, O, O, O, O, O, O, O, O, O, O, O, O, O, O, P, P, P, P, P, P, P, P, P, P, Q, Q, Q, Q, Q, Q, Q, Q, Q, Q, Q, Q, R, R, R, R, R, R, R, R, R, R, R, S, S, S, S, S, S, S, S, S, S, S, S, T, T, T, T, T, T, T, T, T, T, T, T, T, T, U, U, U, U, U, U, U, U, V, V, V, V, V, V, V, V, V, V, V, W, W, W, W, W, W, W, W, W, W, W, W, X, X, X, X, X, X, X, X, X, X, X, X, X, Y, Y, Y, Y, Y, Y, Y, Y, Y, Y, Y, Y, Y, Z, Z, Z, Z, Z, Z, Z, Z, Z, Z, Z, Z, Z, Z
    \item 1, 1, 1, 1, 1, 1, 2, 2, 2, 2, 2, 2, 2, 2, 3, 3, 3, 3, 3, 3, 3, 3, 3, 3, 3, 3, 3, 3, 4, 4, 4, 4, 4, 4, 4, 4, 4, 5, 5, 5, 5, 5, 5, 5, 5, 6, 6, 6, 6, 6, 6, 6, 6, 6, 6, 7, 7, 7, 7, 7, 7, 7, 7, 7, 7, 8, 8, 8, 8, 8, 8, 8, 8, 8, 8, 8, 8, 8, 8, 9, 9, 9, 9, 9, 9, 9, 9, 9, 9, 9, A, A, A, A, A, A, A, A, A, A, A, A, B, B, B, B, B, B, B, B, B, B, B, B, B, C, C, C, C, C, C, C, C, C, C, C, C, C, C, C, C, C, C, C, D, D, D, D, D, D, D, D, D, D, E, E, E, E, E, E, E, E, E, E, E, F, F, F, F, F, F, F, F, F, F, F, F, F, F, F, F, F, F, G, G, G, G, G, G, G, G, G, G, G, H, H, H, H, H, H, H, H, H, H, H, H, H, I, I, I, I, I, I, I, I, J, J, J, J, J, J, J, J, J, K, K, K, K, K, K, L, L, L, L, L, Y, L, L, L, L, L, L, L, L, L, M, M, M, M, M, M, M, M, M, M, M, N, N, N, N, N, N, N, O, O, O, O, O, O, O, O, O, O, O, O, O, O, O, O, O, P, P, P, P, P, P, P, P, P, P, Q, Q, Q, Q, Q, Q, Q, Q, Q, Q, Q, Q, R, R, R, R, R, R, R, R, R, R, R, S, S, S, S, S, S, S, S, S, S, S, S, T, T, T, T, T, T, T, T, T, T, T, T, T, T, U, U, U, U, U, U, U, U, V, V, V, V, V, V, V, V, V, V, V, W, W, W, W, W, W, W, W, W, W, W, W, X, X, X, X, X, X, X, X, X, X, X, X, X, Y, L, Y, Y, Y, Y, Y, Y, Y, Y, Y, Y, Y, Z, Z, Z, Z, Z, Z, Z, Z, Z, Z, Z, Z, Z, Z
    \item \textbf{1, 1, 1, 1, 1, 1, 2, 2, 2, 2, 2, 2, 2, 2, 3, 3, 3, 3, 3, 3, 3, 3, 3, 3, 3, 3, 3, 3, 4, 4, 4, 4, 4, 4, 4, 4, 4, 5, 5, 5, 5, 5, 5, 5, 5, 6, 6, 6, 6, 6, 6, 6, 6, 6, 6, 7, 7, 7, 7, 7, 7, 7, 7, 7, 7, 8, 8, 8, 8, 8, 8, 8, 8, 8, 8, 8, 8, 8, 8, 9, 9, 9, 9, 9, 9, 9, 9, 9, 9, 9, A, A, A, A, A, A, A, A, A, A, A, A, B, B, B, B, B, B, B, B, B, B, B, B, B, C, C, C, C, C, C, C, C, C, C, C, C, C, C, C, C, C, C, C, D, D, D, D, D, D, D, D, D, D, E, E, E, E, E, E, E, E, E, E, E, F, F, F, F, F, F, F, F, F, F, F, F, F, F, F, F, F, F, G, G, G, G, G, G, G, G, G, G, G, H, H, H, H, H, H, H, H, H, H, H, H, H, I, I, I, I, I, I, I, I, J, J, J, J, J, J, J, J, J, K, K, K, K, K, K, L, L, L, L, L, L, L, L, L, L, L, L, L, L, L, M, M, M, M, M, M, M, M, M, M, M, N, N, N, N, N, N, N, O, O, O, O, O, O, O, O, O, O, O, O, O, O, O, O, O, P, P, P, P, P, P, P, P, P, P, Q, Q, Q, Q, Q, Q, Q, Q, Q, Q, Q, Q, R, R, R, R, R, R, R, R, R, R, R, S, S, S, S, S, S, S, S, S, S, S, S, T, T, T, T, T, T, T, T, T, T, T, T, T, T, U, U, U, U, U, U, U, U, V, V, V, V, V, V, V, V, V, V, V, W, W, W, W, W, W, W, W, W, W, W, W, X, X, X, X, X, X, X, X, X, X, X, X, X, Y, Y, Y, Y, Y, Y, Y, Y, Y, Y, Y, Y, Y, Z, Z, Z, Z, Z, Z, Z, Z, Z, Z, Z, Z, Z, Z \textcolor{darkgreen}{(GT)}}
\end{enumerate}
}

\clearpage 

\subsection*{\Large III. Visual Reasoning}

\subsubsection*{\large AIQ Pattern Completion}

\noindent\textbf{Example A. (easy item):} \textit{“Which of the response options (A, B, C, D, or E) can replace the question mark (?) to complete the puzzle?”}

\begin{figure}[htbp]
    \centering
    \includegraphics[width=0.7\textwidth]{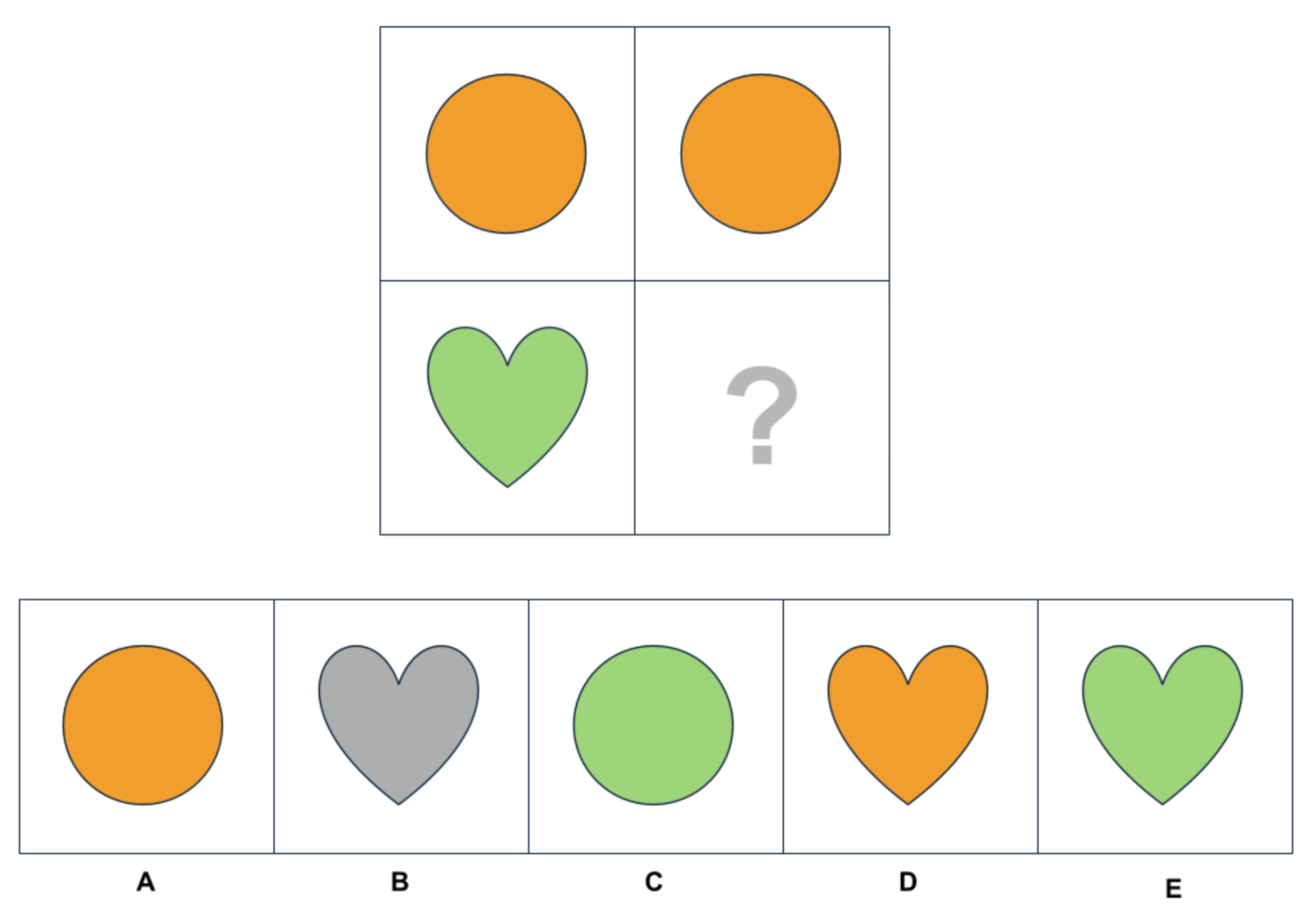}
\end{figure}

\begin{itemize}
    \item[] \textbf{GT: Answer E}
\end{itemize}

\noindent\rule{\textwidth}{0.4pt}

\vspace{0.4cm}

\noindent\textbf{Example B. (difficult item):} \textit{“Which of the response options (A, B, C, D, or E) can replace the question mark (?) to complete the puzzle?”}

\begin{figure}[H]
    \centering
    \includegraphics[width=0.9\textwidth]{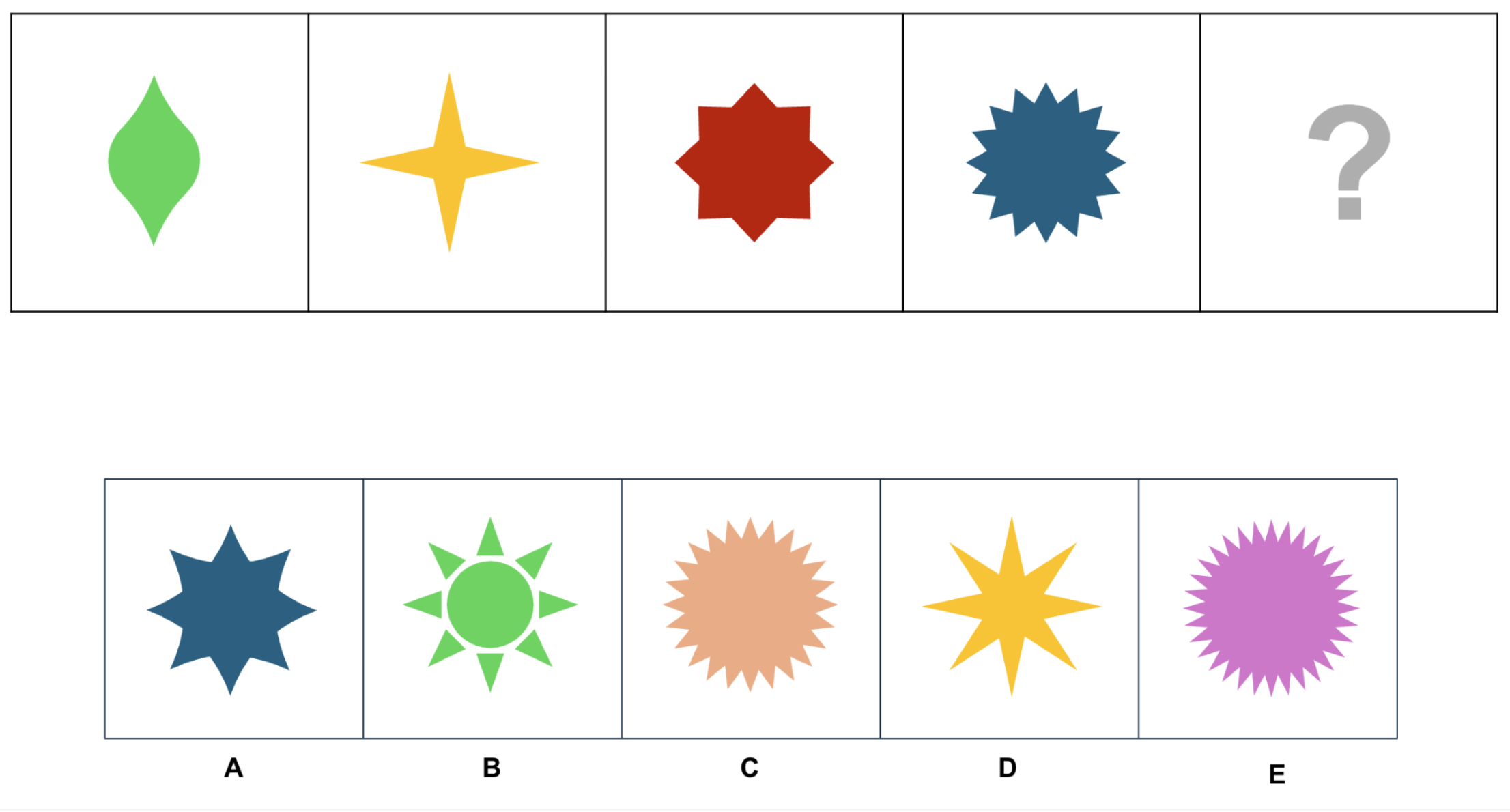}
\end{figure}

\begin{itemize}
    \item[] \textbf{GT: Answer E}
\end{itemize}


\subsubsection*{\large AIQ Algebraic Reasoning (image)}

\noindent\textbf{Example A. (easy item):} \textit{Which one of the response items A, B, C, D, or E can replace the question mark (?) to balance the scale?}

\begin{figure}[H]
    \centering
    \includegraphics[width=0.7\textwidth]{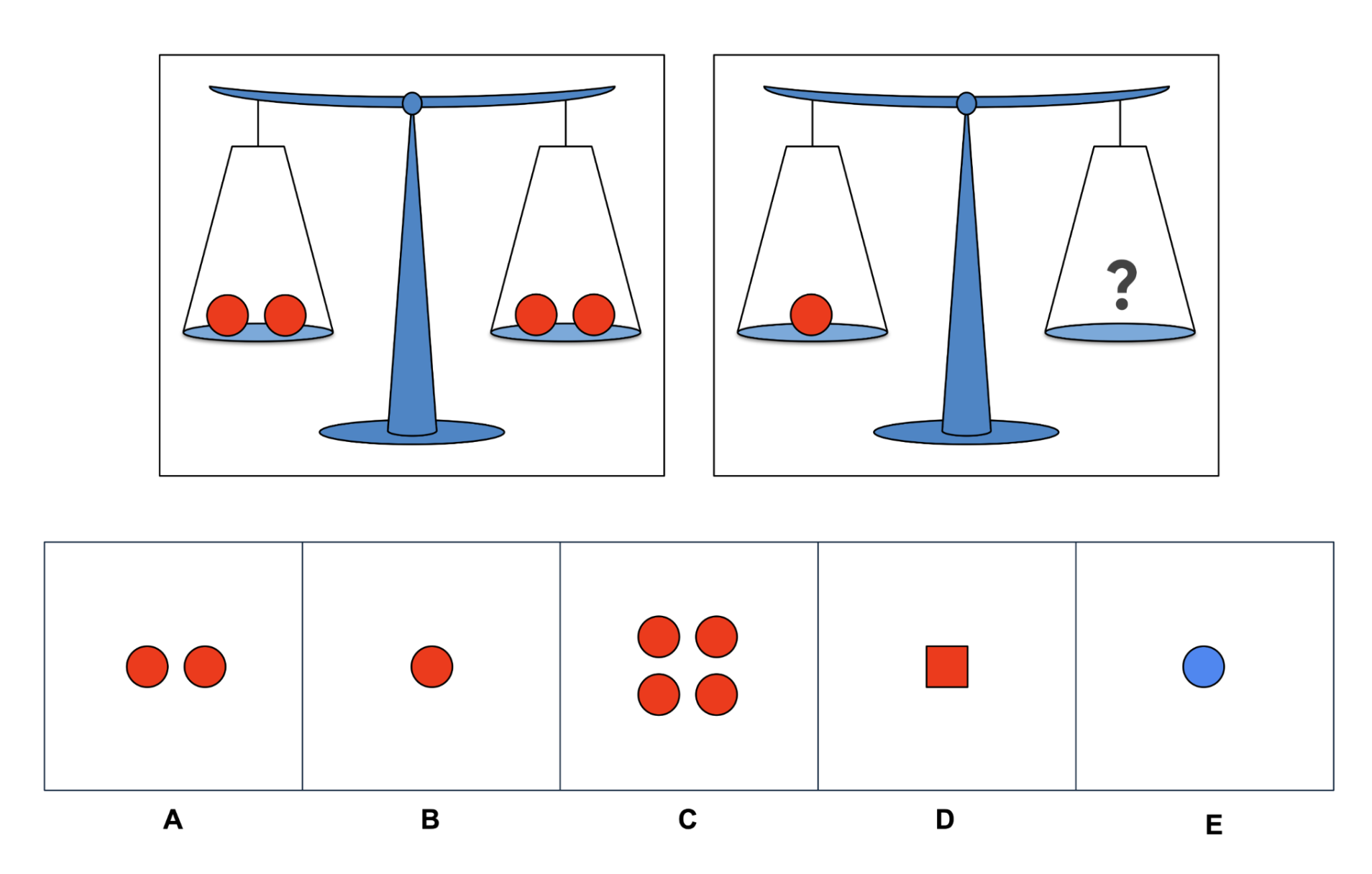}
\end{figure}

\begin{itemize}
    \item[] \textbf{GT: Answer B}
\end{itemize}

\noindent\rule{\textwidth}{0.4pt}

\vspace{.7cm}

\noindent\textbf{Example B. (difficult item):} \textit{Now you have to look at all four scales to find the right answer. Which one of the response items A, B, C, D, or E can replace the question mark (?) to balance the scale?}

\begin{figure}[H]
    \includegraphics[width=1\textwidth]{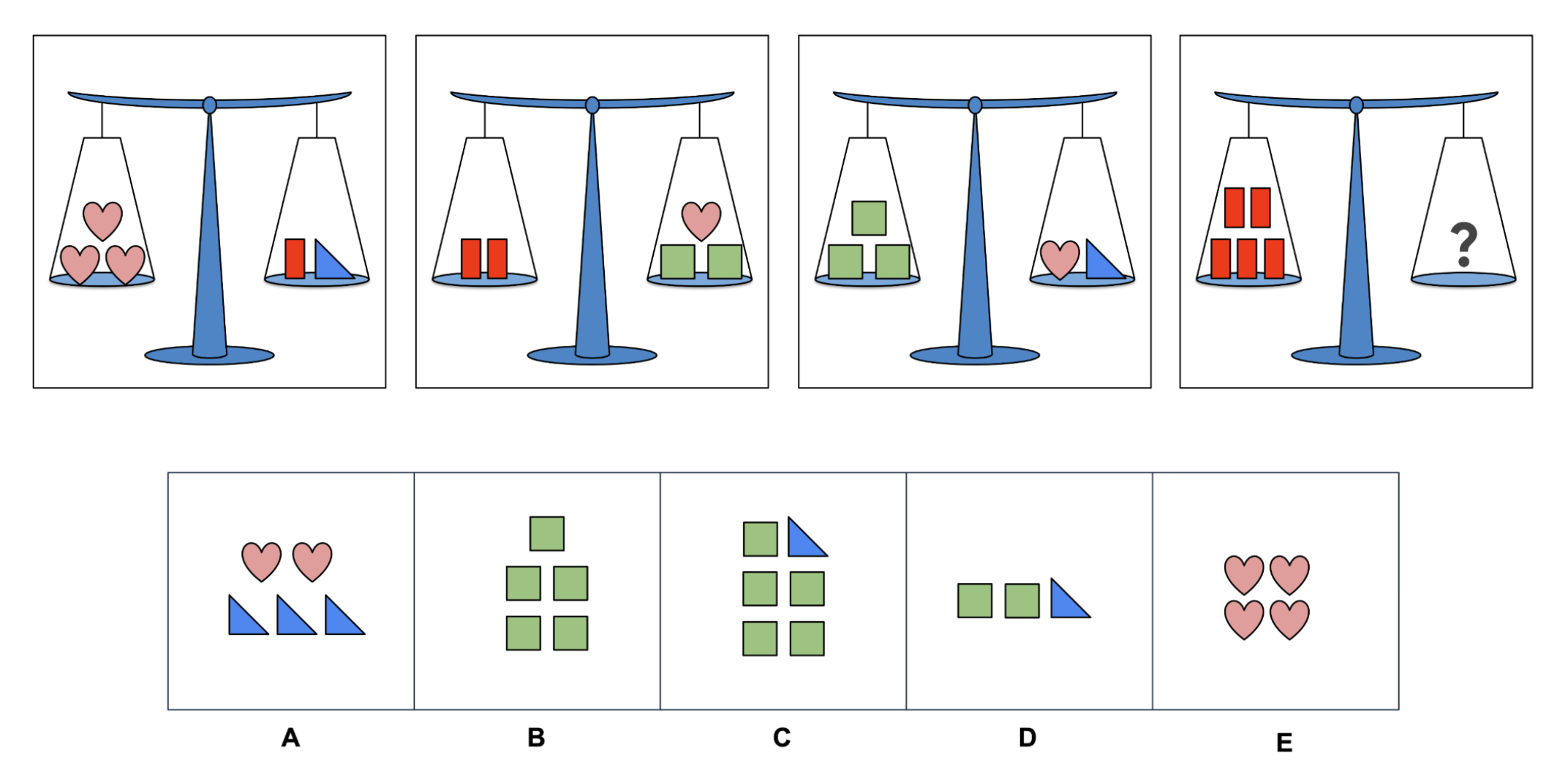} 
\end{figure}

\begin{itemize}
    \item[] \textbf{GT: Answer A}
\end{itemize}


\subsubsection*{\large AIQ Anomaly Detection}

\noindent\textbf{Example A (easy item):} \textit{Which of the following response options (A, B, C, or D) correctly identifies what is missing in this picture?}
\vspace{0.1cm}

\noindent
\begin{minipage}[c]{0.3\textwidth}
    \begin{enumerate}[label=\Alph*., leftmargin=3em]
        \item \textbf{Nose \textcolor{darkgreen}{(GT)}}
        \item Lips
        \item Sweater
        \item Hair
    \end{enumerate}
\end{minipage}%
\hfill
\begin{minipage}[c]{0.6\textwidth}
    \centering
    \includegraphics[width=\linewidth]{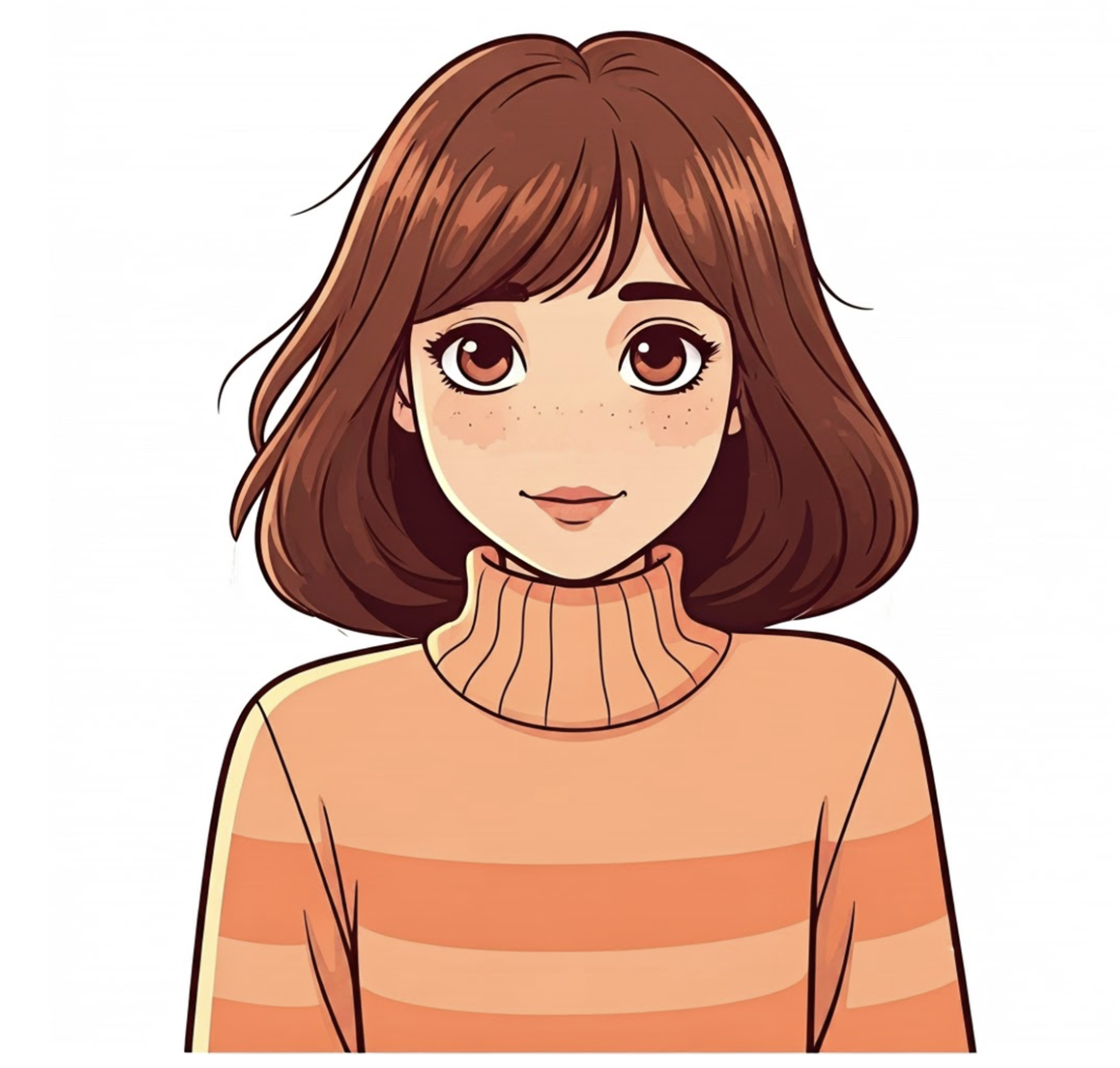}
\end{minipage}

\noindent\rule{\textwidth}{0.4pt}
\vspace{0.1cm} 

\noindent\textbf{Example B (difficult item):} \textit{Which of the following response options (A, B, C, or D) correctly identifies what is missing in this picture?}
\vspace{0cm}

\noindent
\begin{minipage}[c]{0.3\textwidth}
    \begin{enumerate}[label=\Alph*., leftmargin=3em]
        \item Shoe
        \item Shirt Sleeve
        \item \textbf{Backpack straps \textcolor{darkgreen}{(GT)}}
        \item Hair
    \end{enumerate}
\end{minipage}%
\hfill
\begin{minipage}[c]{0.62\textwidth}
    \centering
    \includegraphics[width=\linewidth]{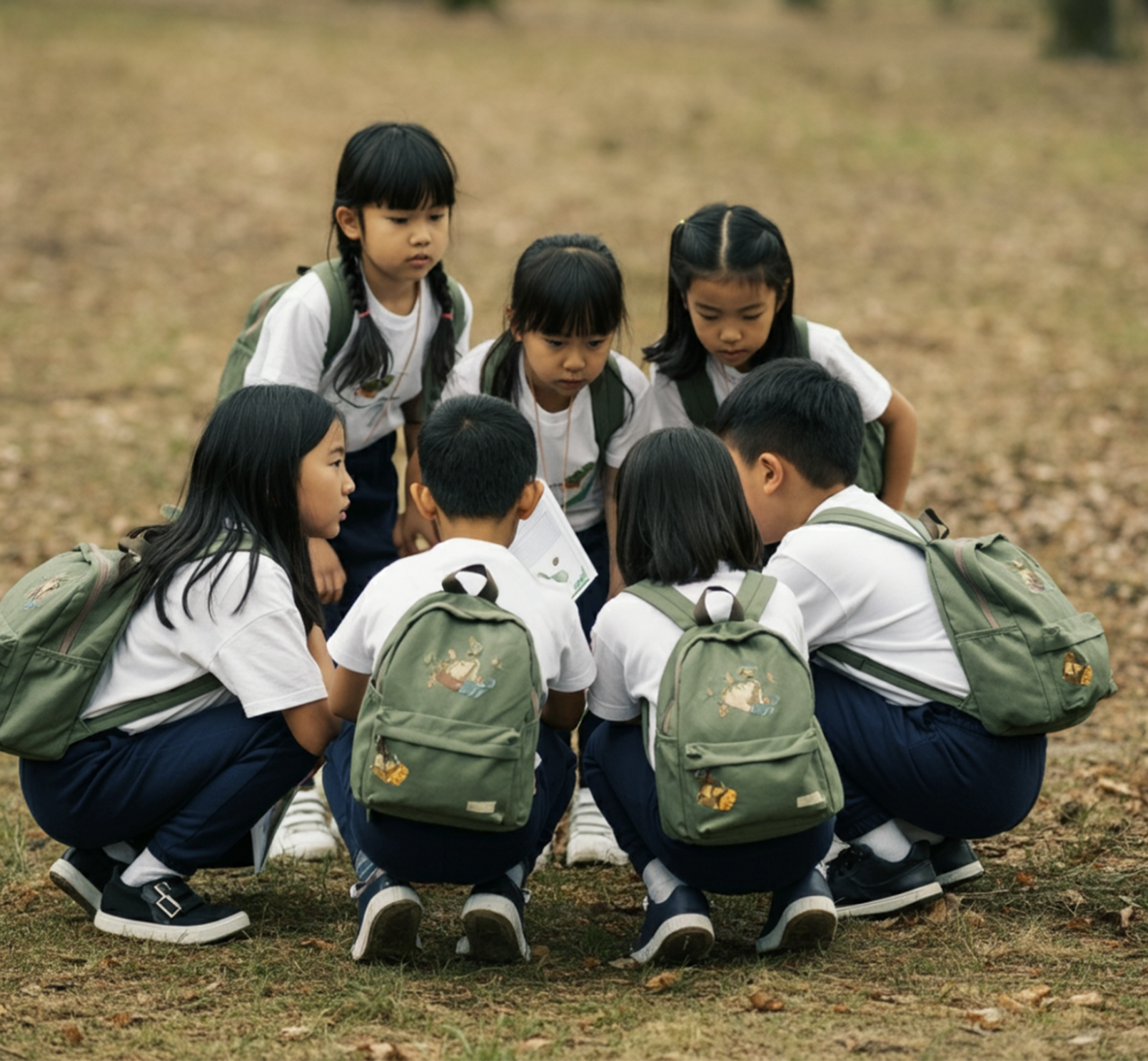}
\end{minipage}%

\captionsetup{font=bf}

\end{document}